\documentclass[final]{cvpr}

\usepackage{times}
\usepackage{epsfig}
\usepackage{graphicx}
\usepackage{amsmath}
\usepackage{amssymb}

\usepackage{adjustbox}
\usepackage{subfig}
\usepackage{comment}
\usepackage{color}

\usepackage{booktabs}
\usepackage{algorithm}
\usepackage{algorithmic}
\usepackage{multirow}

\usepackage{eqparbox}

\usepackage{etoolbox}  % patch def of algorithmic environment
\makeatletter
\patchcmd{\algorithmic}{\addtolength{\ALC@tlm}{\leftmargin} }{\addtolength{\ALC@tlm}{\leftmargin}}{}{}
\makeatother

\usepackage{amsmath}
\DeclareMathOperator*{\argmin}{argmin} % thin space, limits underneath in displays

\newcommand{\pnp}{P\textit{n}P}

\newcommand{\ccar}{\textit{Car}}
\newcommand{\cvanbus}{\textit{Van-Bus}}
\newcommand{\cpickup}{\textit{PickupTruck}}
\newcommand{\cbox}{\textit{BoxTruck}}

\usepackage[pagebackref=true,breaklinks=true,colorlinks,bookmarks=false]{hyperref}

\def\cvprPaperID{7} % *** Enter the CVPR Paper ID here
\def\confYear{CVPR 2021}

\begin{document}

%%%%%%%%% TITLE
\title{1-Point RANSAC-Based Method for Ground Object Pose Estimation}

\author{Jeong-Kyun Lee\textsuperscript{1}, Young-Ki Baik\textsuperscript{1}, Hankyu Cho\textsuperscript{2}\textsuperscript{*}, Kang Kim\textsuperscript{2}\thanks{This work was done while the authors were with Qualcomm.} , Duck Hoon Kim\textsuperscript{1} \\
\textsuperscript{1}Qualcomm Korea YH, Seoul, South Korea \\
\textsuperscript{2}XL8 Inc., Seoul, South Korea
%{\tt\small \{ljeongky,ybaik,duckhoon\}@qti.qualcomm.com}
% For a paper whose authors are all at the same institution,
% omit the following lines up until the closing ``}''.
% Additional authors and addresses can be added with ``\and'',
% just like the second author.
% To save space, use either the email address or home page, not both
%\and
%Hankyu Cho\textsuperscript{*}, Kang Kim\thanks{This work was done while the authors were with Qualcomm.} \\ 
%XL8 Inc., Seoul, South Korea \\
%{\tt\small \{matthew,kai\}@xl8.ai}
}

\maketitle

%%%%%%%%% ABSTRACT
\begin{abstract}
Solving Perspective-\textit{n}-Point (\pnp) problems is a traditional way of estimating object poses.
Given outlier-contaminated data, a pose of an object is calculated with \pnp~algorithms of $n = \{3,4\}$ in the RANSAC-based scheme.
However, the computational complexity considerably increases along with $n$ and the high complexity imposes severe strain on devices which should estimate multiple object poses in real time.
In this paper, we propose an efficient method based on 1-point RANSAC for estimating the pose of an object on the ground.
In the proposed method, a pose is calculated with 1-DoF parameterization by using a ground object assumption and a 2D object bounding box as an additional observation, thereby achieving the fastest performance among the RANSAC-based methods.
In addition, since the method suffers from the errors of the additional information, we propose a hierarchical robust estimation method for polishing a rough pose estimate and discovering more inliers in a coarse-to-fine manner.
The experiments in synthetic and real-world datasets demonstrate the superiority of the proposed method.
\end{abstract}

%%%%%%%%% BODY TEXT
\section{Introduction}

% \vspace{-5pt}

Perspective-\textit{n}-point (\pnp) is a classical computer vision problem of finding a 6-DoF pose of a camera or an object given \textit{n} 3D points and their 2D projection points in a calibrated camera~\cite{fischler1981random}.
It has been widely used in finding ego-motion of a camera and reconstructing a scene in 3D~\cite{furukawa2009accurate,mur2017orb} as well as estimating a pose and a shape of a known or arbitrary object~\cite{murthy2017recon,murthy2017shape,zeeshan2013detailed}.

Handling outliers is a crucial ability in practical applications of pose estimation because mislocalization errors or mismatches of point correspondences inevitably arise.
However, a majority of studies on the \pnp~problem focus on producing high accuracy in noisy data.
Instead, they depend on the RANSAC-based scheme~\cite{fischler1981random} to handle data contaminated with outliers.
The common scheme that \pnp~algorithms with $n=\{3,4\}$ (\ie, P3P~\cite{gao2003complete,kneip2011novel,nister2007minimal} or P4P~\cite{bujnak2008general,horaud1989analytic}) are incorporated into RANSAC distinguishes inlier 2D-3D point correspondences and produces a rough pose estimate which is subsequently polished using the inlier set.
Although the stopping criterion of RANSAC iteration is adaptively determined during the RANSAC process, the average number of trials in this scheme exponentially increases along with \textit{n} in the data with a high outlier ratio.
The high complexity may hinder the system from running on real-world applications, such as localizing multiple objects at once, where fast and accurate execution is crucial.

Ferraz et al. proposed REP\pnp~\cite{ferraz2014very} that is similar to the popular robust estimation technique~\cite{huber2004robust} with iterative re-weighting mechanism.
It estimates a pose by calculating the null space of a linear system for control points in the same way of E\pnp~\cite{lepetit2009epnp}.
Afterward, in an iterative fashion, it assigns confidence values to the correspondences by computing algebraic errors and computes the null space of the weighted linear system. 
Since the repetition is empirically finished within several times, REP\pnp~achieves fast and accurate performance.
However, as with a general M-estimator~\cite{huber2004robust}, its highly possible breakdown point can attain 50\%.
Thus, this method does not ensure working on the data with the high outlier ratio.

In this paper, we propose an efficient method for calculating the 6D pose of an object with a P1P~algorithm in the RANSAC-based scheme.
To simplify the 6D pose estimation problem, we assume an object is on the ground and the relationship between the ground and the camera is pre-calibrated.
Then, given a 2D object bounding box and $n$ 2D-3D point correspondences, the \pnp~problem is reformulated by 1-DoF parameterization (\ie, the yaw angle or depth of an object), thereby producing a pose estimate from one point correspondence.
The synthetic experiments demonstrate our method is the fastest one among RANSAC-based methods.
However, the proposed method suffers from erroneous 2D bounding box or ground geometry.
Therefore, we also propose a hierarchical robust estimation method for improving the performance practically.
In the refinement stage, it not only polishes the rough pose estimate but also secures more inliers in a coarse-to-fine manner.
It consequently achieves more robust and accurate performance in the erroneous cases and a more complex problem (\eg, joint pose and shape estimation~\cite{murthy2017recon,murthy2017shape,zeeshan2013detailed}).

\vspace{-2pt}

% ============== Related works
\section{Related Works} \label{sec:relatedwork}

\vspace{-5pt}

% Outlier data
As mentioned above, the \pnp~problem in the data containing outliers is handled with REP\pnp~\cite{ferraz2014very} or usually the RANSAC scheme~\cite{fischler1981random} where P3P~\cite{gao2003complete,kneip2011novel,nister2007minimal} or P4P~\cite{bujnak2008general,horaud1989analytic} are incorporated for providing minimal solutions of the \pnp~problem.
% Intro of PnP
Except for them, the existing \pnp~algorithms have aimed at improving the performance in noisy data.
The \pnp~algorithms are traditionally categorized into two types of methodologies: iterative and non-iterative methods.

% iterative methods
Iterative methods~\cite{dementhon1995model,garro2012solv,horaud1997object,lowe1991fitting,lu2000fast} find a globally optimal pose estimate by solving a nonlinear least squares problem to minimize algebraic or geometric errors with iterative or optimization techniques.
Among them, the method proposed by Lu~\etal~\cite{lu2000fast} accomplishes outstanding accuracy.
It reformulates an objective function as minimizing object-space collinearity errors instead of geometric errors measured on the image plane.
The least squares problem is solved using the way of Horn~\etal~\cite{horn1988closed} iteratively.
Garro~\etal~\cite{garro2012solv} proposed a Procrustes \pnp~(P\pnp) method reformulating the \pnp~problem as an anisotropic orthogonal Procrustes (OP) problem~\cite{dosse2011anisotropic}.
They iteratively computed a solution of the OP problem by minimizing the geometric errors in the object space until convergence, which achieved a proper trade-off between speed and accuracy.

% Non-iterative methods
On the other hand, non-iterative methods~\cite{hesch2011direct,kneip2014upnp,lepetit2009epnp,li2012robust,zheng2013revisiting,zheng2013aspnp} quickly obtain pose estimates by calculating solutions of a closed form.
% EPnP
Lepetit~\etal~\cite{lepetit2009epnp} proposed an efficient and accurate method (E\pnp) for solving the \pnp~problem with computational complexity of $O(n)$.
They defined the \pnp~problem as finding virtual control points, which was quickly calculated by null space estimation of a linear system.
They refined the solution with the Gauss-Newton method so that its accuracy amounted to that of Lu~\etal~\cite{lu2000fast} with less computational time.
% DLS
Hesch~\etal~\cite{hesch2011direct} formulated the \pnp~problem as finding a Cayley-Gibbs-Rodrigues (CGR) parameter by solving a system of several third-order polynomials.
% ASPnP, OPnP, UPnP
Several methods~\cite{kneip2014upnp,zheng2013revisiting,zheng2013aspnp} employ quaternion parameterization and solve a polynomial system minimizing algebraic~\cite{zheng2013revisiting,zheng2013aspnp} or object-space~\cite{kneip2014upnp} errors with the Gröbner basis technique~\cite{kukelova2008automatic}.
In particular, the method of Kneip~\etal~\cite{kneip2014upnp} results in minimal solutions of the polynomial system and is generalized for working on both central and non-central cameras.
% RPnP
Li~\etal~\cite{li2012robust} proposed a \pnp~approach robust to special cases such as planar and quasi singular cases.
They defined the \pnp~problem by finding a rotational axis, an angle, and a translation vector, and solved the linear systems formulated from projection equations and a series of 3-point constraints~\cite{quan1999linear}.
% CEPPnP/MLPnP - Confidence / weighted methods
Recently, some methods~\cite{ferraz2014leveraging,urban2016mlpnp} utilize the uncertainty of observations on the image space.
They estimate control points~\cite{ferraz2014leveraging} or directly a pose parameter~\cite{urban2016mlpnp} by performing maximum likelihood estimation for Sampson errors reflecting covariance information of observations.

\vspace{-2pt}

% ============== Proposed Method
\section{Proposed Method}

\vspace{-5pt}

In this paper, we handle how to reduce the computational complexity of a RANSAC-based scheme for the \pnp~problem in outlier-contaminated data.
Many real-world applications~\cite{murthy2017recon,murthy2017shape,zeeshan2013detailed} aim at estimating a 6D pose of an object on the ground, given 2D object bounding boxes from object detectors~\cite{girshick2015fast,liu2016ssd}.
The additional information is not only useful to reduce the number of required points for solving the \pnp~problem but also acquired without an extra computational loss since it is already given in some applications.
%reasonable due to being acquired without an extra computational loss in some applications.
Hence, we propose a perspective-1-point (P1P) method for an object on the ground, which significantly raises the speed of the RANSAC-based scheme.
In the following sections, we first introduce a general framework for $n$-point RANSAC-based pose estimation, then propose the P1P method for roughly estimating an object pose using one point sample, and finally suggest a novel robust estimation method for polishing the rough pose estimate.

\vspace{-2pt}

% ============== 1-point RANSAC-based method: A general framework of RANSAC-based pose estimation
\subsection{General framework of $n$-point RANSAC-based pose estimation}

\vspace{-5pt}

\begin{figure}[t]
	\begin{center}
		\includegraphics[width=.89\linewidth]{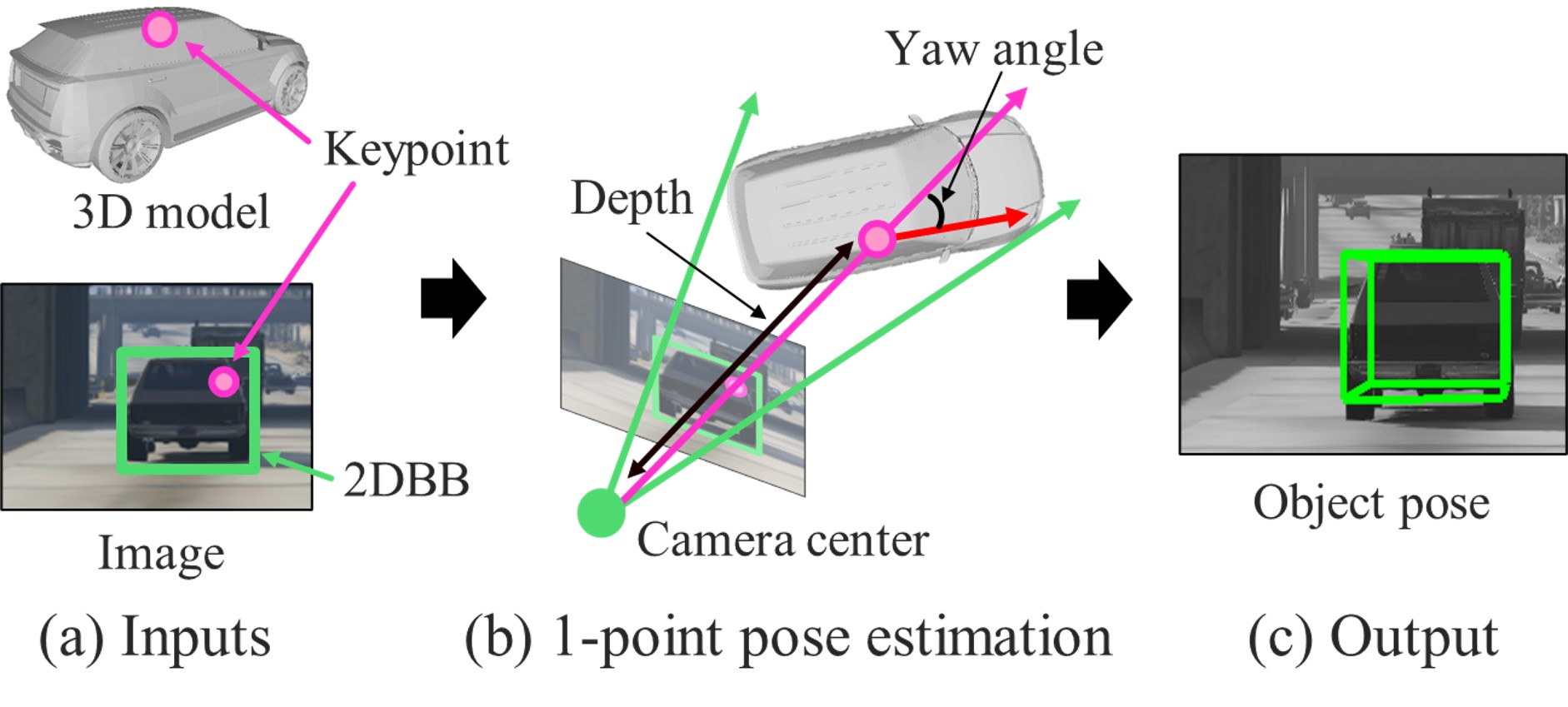} \vspace{-15pt}
	\end{center}
	\caption{A flow chart of the proposed method for pose estimation} \vspace{-13pt}
	\label{fig:overall} 
\end{figure}

We employ a general framework for $n$-point RANSAC-based pose estimation.\footnote{Please find a flowchart of the RANSAC-based scheme in the supplementary material.}
Given a set of 2D-3D keypoint correspondences, $\mathcal{C}$, we sample $n$ keypoint correspondences, \ie,  $\left\{ (\mathbf{x}_1,\mathbf{X}_1), ..., (\mathbf{x}_n,\mathbf{X}_n) | (\mathbf{x}_i,\mathbf{X}_i)\in \mathcal{C} \right\}$, and compute a pose candidate $\mathbf{T}_{cand}$ using a P$n$P algorithm.
Then the reprojection errors are computed for all the keypoint correspondences and the keypoints whose reprojection errors are within a threshold $t_{in}$ are regarded as inliers.
This process is repeated $N$ times and we select the pose with the maximum number of inliers as the best pose estimate.
The maximum number of iteration, $N$, can be reduced by the adaptive RANSAC technique~\cite{fischler1981random} which adjusts $N$ depending on the number of inliers during the RANSAC process.
Finally, the pose estimate is polished by minimizing the reprojection errors of the inlier keypoints or using the existing P$n$P algorithms.

\vspace{-5pt}

% ============== 1-point RANSAC-based method: 1-point RANSAC-based pose estimation
\subsection{Perspective-1-point solution} \label{sec:p1p}

\vspace{-5pt}

To reduce the number of points required for computing an object pose hypothesis, we need to use some prior knowledge.
First, we assume that the tilt of a camera to the ground where an object of interest is placed is given as a pitch angle $\theta_{p}$.\footnote{It is pre-calibrated or calculated in an online manner with \cite{jeong2016adaptive,zhang2014robust}.}
From the pitch angle, a rotation transformation $\mathbf{R}_{cg} \in SO(3)$ from the ground to the camera is defined as $\mathbf{R}_{cg} = e^{\mathbf{\omega}_{p}}$ where $\mathbf{\omega}_{p} = [ -\theta_{p}, 0, 0 ]^\top \in so(3)$ and $e: so(3) \mapsto SO(3)$ is an exponential map.
For example, if an object is fronto-parallel to the image plane, $\mathbf{R}_{cg} = \mathbf{I}_3$ where $\mathbf{I}_3$ is a $3\times3$ identity matrix.
Second, besides the 3D model of the object and its 2D-3D keypoint correspondence, which are provided as input by default in the P$n$P problem, we assume that the 2D bounding box (2DBB) of the object of interest in an image is given as an input.
Using the additional prior information, the problem is redefined as aligning the projection of the 3D bounding box (3DBB) into the back-projected rays of both the side ends of its 2DBB in a bird-eye view (BEV), thereby computing the yaw angle of the object and the depth of the keypoint as shown in Figure~\ref{fig:overall}b.
Here, we formulate an equation that computes the pose of the object of interest by 1-DoF parameterization.
A unique pose hypothesis per keypoint is obtained by solving the equation.
Consequently, through the RANSAC process, we find the best pose parameter with the maximum number of inliers among these hypotheses and then optimize the pose.

\begin{figure}[t]
	\begin{center}
	\subfloat[]{
	    \adjustbox{raise=-2.3pc}{
	        \includegraphics[width=.17\linewidth,height=65pt]{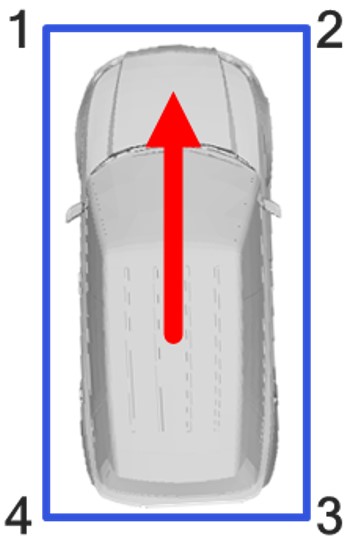}
	        \label{fig:4cases_a}
	        }
	}
	\subfloat[]{
	    {\setlength\tabcolsep{0.3ex}
	    \small
	    \begin{tabular}{cccc}	    
            \includegraphics[width=.17\linewidth]{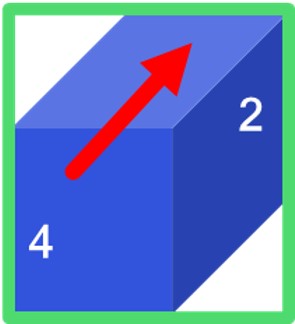} &
	        \includegraphics[width=.17\linewidth]{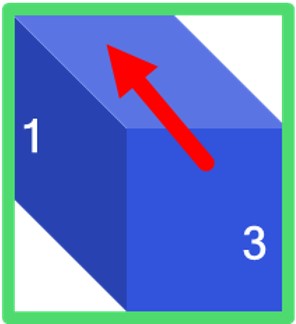} &
	        \includegraphics[width=.17\linewidth]{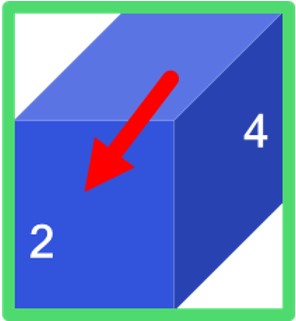} &
	        \includegraphics[width=.17\linewidth]{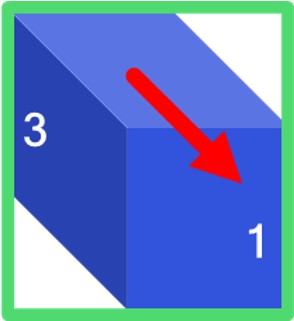} \\
	        Case 1 & Case 2 & Case 3 &  Case 4 
	    \end{tabular}
	    \label{fig:4cases_b}
	    }
	}
	\end{center} \vspace{-14pt}
	\caption{Case study on the 1-point object pose estimation: (a) 3DBB of an object in the BEV and (b) four cases of the pose estimation depending on the yaw angle} \vspace{-11pt}
	\label{fig:4cases} 
\end{figure}

\begin{figure}[t]
	\begin{center}
	\subfloat[]{
	    \adjustbox{raise=-4.05pc}{
	        \includegraphics[width=.405\linewidth]{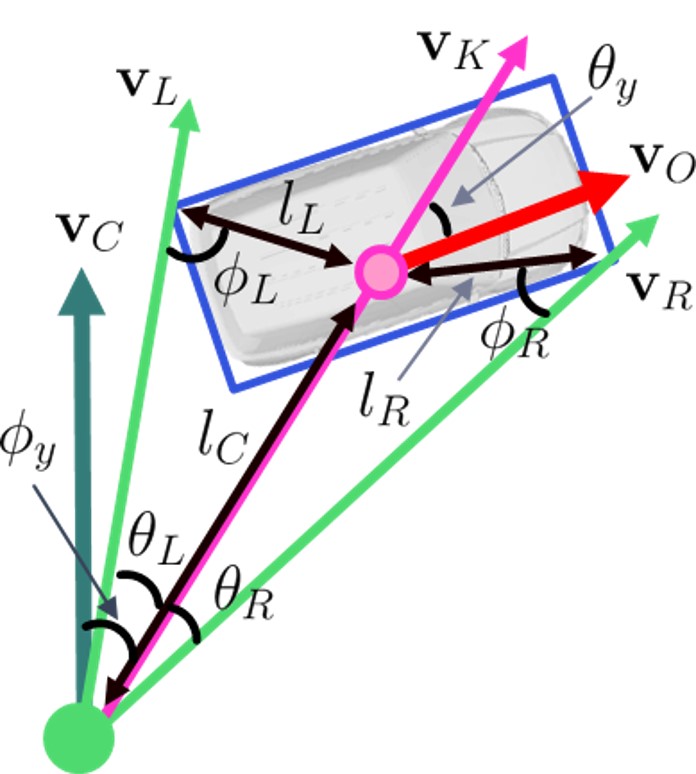}
	        \label{fig:params_a}
        }
	}
	\subfloat[]{
	    {\setlength\tabcolsep{0.3ex}
	    \small
	    \begin{tabular}{cc}
            \includegraphics[width=.245\linewidth,height=35pt]{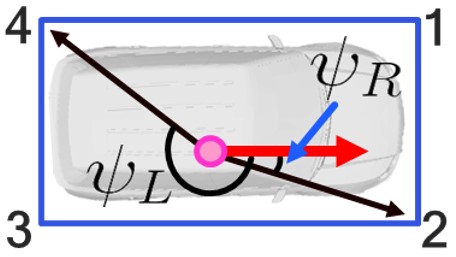} &
	        \includegraphics[width=.245\linewidth,height=35pt]{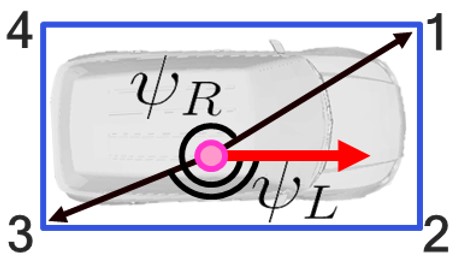} \\
	        Case 1 & Case 2 \\[1ex]
	        \includegraphics[width=.245\linewidth,height=35pt]{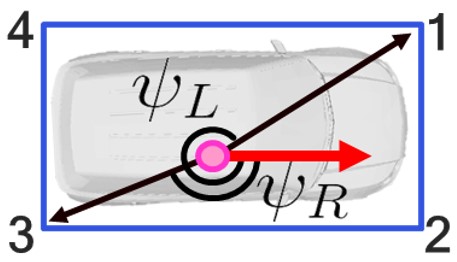} &
	        \includegraphics[width=.245\linewidth,height=35pt]{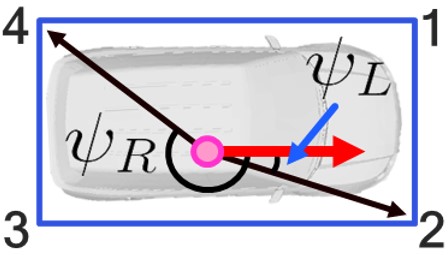} \\
	        Case 3 &  Case 4
	    \end{tabular}
	    \label{fig:params_b}
	    }
	}
	\end{center} \vspace{-14pt}
	\caption{Parameter definition in the BEV: (a) parameters in the camera coordinate system and (b) parameters for each case in the 3D model coordinate system} \vspace{-11pt}
	\label{fig:params} 
\end{figure}

There are four cases of the 1-point object pose estimation depending on the yaw angle of the object as shown in Figure~\ref{fig:4cases}.
In Figure~\ref{fig:4cases_a}, the red arrow and the blue rectangle denote the forward direction of the object and the 3DBB in the BEV, respectively.
We denote the four corners of the 3DBB by numbers which will be used in Figure~\ref{fig:params_b}.
In Figure~\ref{fig:4cases_b}, the green rectangle and the blue hexahedron denote the 2DBB of the object and the projection of the 3DBB to an image, respectively.
Then, as shown in Figure~\ref{fig:4cases_b}, the left and right edges of the 2DBB are adjacent to two of the corners of the 3DBB depending on the yaw angle of the object, \eg, in Case 1, the left and right edges of the 2DBB are paired with Corners 4 and 2 of the 3DBB, respectively.
In this way, we can find out the four cases\footnote{In fact, the number of the cases are exactly more than 4 due to the perspective effect, \textit{e.g.}, Corners 3 and 4 of the 3DBB are possible to be adjacent to the 2DBB, but those cases can approximate to one of the cases in Figure~\ref{fig:4cases_b} if an object of interest is far enough from the camera.} for the 1-point pose estimation and will describe how to compute an object pose using a point correspondence at each case.

Figure~\ref{fig:params} shows the definition of the parameters required to formulate an equation in the BEV.
In Figure~\ref{fig:params_a}, $\mathbf{v}_K$ (magenta) is a directional vector from the camera center to the keypoint, $\mathbf{v}_L$ and $\mathbf{v}_R$ (green) are the back-projected rays of the left and right edges of the 2DBB\footnote{We assume that the side edges of the 2DBB are back-projected to the planes perpendicular to the ground so the projections of the planes into the ground become the rays passing through the camera center and the corners of the 3DBB.}, respectively, $\mathbf{v}_O$ (red) is the forward direction of the object, $\mathbf{v}_C$ (dark green) is the forward direction of the camera, $\phi_y$, $\theta_L$, and $\theta_R$ are angles between $\mathbf{v}_C$ and $\mathbf{v}_K$, $\mathbf{v}_K$ and $\mathbf{v}_L$, and $\mathbf{v}_K$ and $\mathbf{v}_R$, respectively, $l_C$ is the length between the camera center point and the keypoint, $l_L$ and $l_R$ are the lengths between the keypoint and the corners of the 3DBB, and $\phi_L$ and $\phi_R$ are angles between $\mathbf{v}_L$ and $l_L$, and $\mathbf{v}_R$ and $l_R$, respectively.
In Figure~\ref{fig:params_b}, $\psi_L$ and $\psi_R$ are angles between the forward direction of the object and the corners of the 3DBB.
Then our goal is to compute a local yaw angle of the object, \ie, an angle between $\mathbf{v}_K$ and $\mathbf{v}_O$, $\theta_y$, and a depth from the camera center to the keypoint, $l_C$.

We first describe the 1-point pose estimation method for Case 1.
We can derive an equation to compute $\theta_y$ from the sine rule as follows:
\begin{equation} \label{eq:sinrule}
    l_C = \frac{l_R \sin{\phi_R}}{\sin{\theta_R}} = \frac{l_L \sin{\phi_L}}{\sin{\theta_L}},
\end{equation}
where $\phi_R = \theta_y + \psi_R - \theta_R$ and $\phi_L = -\theta_y - \psi_L - \theta_L$.
From Eq.~\ref{eq:sinrule}, the yaw angle $\theta_y$ is derived as
\begin{equation} \label{eq:yaw}
    \theta_y = \arctan{\left( \frac{-\frac{l_R \sin{\left(\psi_R - \theta_R\right)}}{\sin{\theta_R}} - \frac{l_L \sin{\left(\psi_L + \theta_L\right)}}{\sin{\theta_L}}}{\frac{l_R \cos{\left(\psi_R - \theta_R\right)}}{\sin{\theta_R}} + \frac{l_L \cos{\left(\psi_L + \theta_L\right)}}{\sin{\theta_L}}} \right)},
\end{equation}
and the depth of the keypoint, $l_C$, is calculated using Eq.~(\ref{eq:sinrule}).\footnote{Please see the supplementary material for the details of the derivation.}

Once $\theta_y$ and $l_C$ are calculated, an object pose $\mathbf{T}_{co} \in SE(3)$, which is a transformation matrix from the object coordinates to the camera coordinates, can be computed by
\begin{equation} \label{eq:Tco}
	\mathbf{T}_{co} = \begin{bmatrix} \mathbf{R}_{cg} & \mathbf{0}_3 \\ \mathbf{0}_3^\top &  1 \end{bmatrix} 
	\begin{bmatrix} e^{\omega_y} & l_C\mathbf{d}_\mathbf{x} \\ \mathbf{0}_3^\top &  1 \end{bmatrix}
	\begin{bmatrix} \mathbf{I}_3 & -\mathbf{X} \\ \mathbf{0}_3^\top &  1 \end{bmatrix},
\end{equation}
where $\omega_y = [0, \phi_y + \theta_y, 0]^\top$, $\mathbf{X}$ is a 3D location of the selected keypoint, and $\mathbf{d}_\mathbf{x}$ is a normalized directional vector by back-projecting the keypoint $\mathbf{x}$ to a 3D ray. Here, $\mathbf{d}_\mathbf{x}$ can be computed by
\begin{equation}
	\mathbf{d}_\mathbf{x} = \frac{\mathbf{R}_{cg}^\top \mathbf{K}^{-1} \hat{\mathbf{x}}}{\sqrt{\hat{\mathbf{x}}^\top \mathbf{K}^{-\top} \mathbf{R}_{cg} \mathbf{S} \mathbf{R}_{cg}^\top \mathbf{K}^{-1} \hat{\mathbf{x}}}}, \ \mathrm{where} \ \mathbf{S} = \begin{bmatrix} 1 & 0 & 0 \\ 0 & 0 & 0 \\ 0 & 0 & 1 \end{bmatrix},
\end{equation}
where $\mathbf{K}$ is a camera intrinsic matrix and $\hat{\mathbf{x}}$ is the homogeneous coordinate of $\mathbf{x}$. 

{
\setlength{\tabcolsep}{4pt}
\renewcommand{\arraystretch}{1.4}
\begin{table}[t] 
    \centering
    \footnotesize
    \caption{Range of angle parameters for the four cases} \vspace{-6pt}
    \begin{tabular}{ccccc} \toprule
         Parameter & Case 1 & Case 2 & Case 3 & Case 4  \\\midrule
         $\theta_y$ & $\left[ 0, \frac{\pi}{2} \right]$ & $\left[ \frac{\pi}{2}, \pi \right]$ & $\left[ -\frac{\pi}{2}, 0 \right]$ & $\left[ -\pi, -\frac{\pi}{2} \right]$ \\
         $\psi_L$ & $\left[ -\pi, -\frac{\pi}{2} \right]$ & $\left[ \frac{\pi}{2}, \pi \right]$ & $\left[ -\frac{\pi}{2}, 0 \right]$ & $\left[ 0, \frac{\pi}{2} \right]$ \\
         $\psi_R$ & $\left[ 0, \frac{\pi}{2} \right]$ & $\left[ -\frac{\pi}{2}, 0 \right]$ & $\left[ \frac{\pi}{2}, \pi \right]$ & $\left[ -\pi, -\frac{\pi}{2} \right]$ \\ \toprule
    \end{tabular}  \vspace{-4pt}
    \label{tab:params}
\end{table} 
}

The object poses for the other cases are similar to Case 1 in terms of computation.
The only difference is the definitions of $\psi_L$ and $\psi_R$ as shown in Figure~\ref{fig:params_b}.
Due to all the cases presented in Figure~\ref{fig:4cases} and the sign inside the arctangent function of Eq.~\ref{eq:yaw}, we have four solutions.
However, as shown in Table~\ref{tab:params}, if we consider the ranges of $\theta_y$, $\psi_L$, and $\psi_R$ for each case, we can filter out unreliable solutions and obtain a unique solution from only one keypoint.

\vspace{-1pt}

% ============== Hierarchical robust estimation
\subsection{Hierarchical robust pose estimation of objects with deformable shapes} \label{sec:hre}

\vspace{-5pt}

Many practical applications~\cite{chhaya2016mono,murthy2017recon,murthy2017shape,zeeshan2013detailed} have dealt with the pose estimation of objects with deformable shapes.
Unlike solving the P$n$P problem using a known 3D model, estimating object pose and shape simultaneously in a single image is an ill-posed problem.
Thus, the existing methods~\cite{chhaya2016mono,murthy2017recon,murthy2017shape,zeeshan2013detailed} have exploited active shape models~(ASM)~\cite{cootes1995active}.
Given a set of class-specified 3D object models as prior information, the $i^{th}$ 3D keypoint location $\mathbf{X}_i$ in the ASM is defined by summing a mean location $\bar{\mathbf{X}}_i$ and the combinations of $M$ basis vectors $\mathbf{B}_i^j$.
\begin{equation}
    \mathbf{X}_i = \bar{\mathbf{X}}_i + \sum^M_j \lambda_j \mathbf{B}_i^j,
\end{equation}
where $\mathbf{\lambda} = \{\lambda_1, ..., \lambda_M\}$ is a set of variables depending on shape variation.

Then the object pose and shape are jointly estimated by minimizing residuals~$r_i$ (reprojection errors) as follows:
\begin{equation} \label{eq:re}
    \argmin_{\mathbf{T},\mathbf{\lambda}} \sum_{i=0}^{|\mathcal{C}|} \rho ( r_i(\mathbf{T}, \lambda) ),
\end{equation}
where
$\ r_i(\mathbf{T},\lambda) = \|f(\mathbf{R} (\bar{\mathbf{X}}_i + \sum^M_{j=0} \lambda_j \mathbf{B}_i^j) + \mathbf{t}) - \mathbf{x}_i \| _2$, 
$|\mathcal{C}|$ is the cardinality of the correspondence set $\mathcal{C}$, the object pose $\mathbf{T}$ is represented by decomposition into a rotation matrix $\mathbf{R}$ and a translation vector $\mathbf{t}$, $f$ is a projection function from a 3D coordinate to an image coordinate, and $\rho$ represents an M-estimator~\cite{huber2004robust} for robust estimation in the presence of outliers.

In our experiments, we use the Tukey biweight function defined by
\begin{equation} \label{eq:tukey}
    \rho(r_i) = \left\{\begin{matrix}
        \frac{c^2}{6} \left \{ 1 - \left [ 1 - \left ( \frac{r_i}{c} \right )^2 \right ]^3 \right \}, & \ \mathrm{if} \ r_i \leq c \ & \mathrm{and} \\ 
        \frac{c^2}{6}, & \ \ \mathrm{if} \ r_i > c. &
    \end{matrix}\right.
\end{equation}
Here, $c = 4.685s$ where a scale factor $s$ is $MAD(r_i)/0.6745$. $MAD(r_i)$ means a median absolute deviation of residuals~$r_i$.
We initialize the coefficients $\lambda$ as zeros and the object pose by the RANSAC process where the unknown 3D model is substituted with the mean shape $\bar{\mathbf{X}}$.
Finally, optimal pose and shape minimizing Eq.~(\ref{eq:re}) are estimated using the iterative reweighted least squares (IRLS) method.

However, the common approach~\cite{lourakis2013model} for robust pose and shape estimation often gets stuck in local minima by the following reasons:
First, the scale factor computed by MAD attains a breakdown point of 50\% in a statistical point of view but does not produce a geometrically meaningful threshold in the data contaminated with outliers.
Second, the M-estimator is sensitive to initial parameters.
In particular, our P1P solution produces a more noisy pose estimate when the camera pitch angle varies or a 2D bounding box from an object detector is erroneous.

Inspired by MM-estimator~\cite{yohai1987high} and annealing M-estimator~\cite{li1998robust} that reduce the sensitivity to the scale estimate and avoid to get stuck in a local minimum via an annealing process, we propose a hierarchical robust estimation method for the object pose and shape estimation.
We repeat M-estimation while decreasing the scale factor.
Here, we use geometrically meaningful and user-defined scale factors because the threshold empirically set by a user may be rather meaningful than one calculated from statistical analysis in the case that camera properties such as intrinsic parameters remain constant and the input data contain outliers.

\begin{algorithm} [t]
	\caption{Hierarchical robust pose and shape estimation}
	\label{alg:hre}
	\small
	\begin{algorithmic}[1]
	    \REQUIRE $\mathcal{C}, \mathbf{T}_{init}, \tau_1, \tau_2, \tau_3$
		\ENSURE $\mathbf{T}, \mathbf{\lambda}$
		\STATE $\mathbf{T} \leftarrow \mathbf{T}_{init}, \lambda \leftarrow \mathbf{0}$
		\STATE $\mathbf{T} \leftarrow \argmin_{\mathbf{T}} \sum^{|\mathcal{C}|}_{i} \rho(r_i(\mathbf{T} | \lambda = 0) | c = 4.685\gamma(s|\tau_2,\tau_3))$
		\STATE $\mathbf{T},\mathbf{\lambda} \leftarrow \argmin_{\mathbf{T},\mathbf{\lambda}} \sum^{|\mathcal{C}|}_{i} \rho(r_i(\mathbf{T},\lambda) | c = 4.685\gamma(s|\tau_1,\tau_2))$
		\STATE $\mathcal{C}_{inlier} \leftarrow \{ (\mathbf{x}_i, \mathbf{X}_i) | r_i(\mathbf{T},\lambda) < \tau_1 \} $
		\STATE $\mathbf{T},\mathbf{\lambda} \leftarrow \argmin_{\mathbf{T},\mathbf{\lambda}} \sum^{|\mathcal{C}_{inlier}|}_{i} \| r_i(\mathbf{T},\lambda) \|^2$
	\end{algorithmic}
\end{algorithm}

The details of the proposed method are described in Algorithm~\ref{alg:hre}.
Given the 2D-3D correspondence set $\mathcal{C}$, an initial pose $\mathbf{T}_{init}$ by the RANSAC process, and user-defined thresholds $\tau_1$, $\tau_2$, and $\tau_3$ ($\tau_1 < \tau_2 < \tau_3$), the object pose $\mathbf{T}$ and shape $\mathbf{\lambda}$ are estimated through three stage optimization.
At the first stage, the roughly initialized pose is refined with the scale estimates loosely bounded to a range of $[\tau_2,\tau_3]$ by the function $\gamma$, which is a clamp function for limiting the range of an input value $x$ to $[\alpha, \beta]$ and defined as $\gamma(x|\alpha,\beta) = \max(\alpha,\min(x,\beta))$.
At the second stage, the pose $\mathbf{T}$ and shape parameter $\lambda$ are jointly optimized with the scale estimates tightly bounded to a range of $[\tau_1,\tau_2]$.
Finally, the pose and shape are polished using only the inlier correspondences set $\mathcal{C}_{inlier}$.
The first and second stages are computed using the IRLS method and the third stage is done using the Gauss-Newton method.
In a case of the P$n$P problem using a known 3D model, the hierarchical robust estimation method can be also used to estimate only an object pose by excluding $\lambda$ in Algorithm~\ref{alg:hre}.

\vspace{-3pt}

% ============== Experiments
\section{Experiments} \vspace{-3pt}

% Synthetic Dataset
\subsection{Synthetic experiments} \label{sec:exp_syn} \vspace{-3pt}

% Dataset
\subsubsection{Dataset}
\vspace{-6pt}
It is assumed that a camera is calibrated with a focal length of 800 pixels and images are captured with resolution of $640\times480$.
Object points are randomly sampled with a uniform distribution in a cube region of $[-2,2]\times[-2,2]\times[-2,2]$.
A center location and a yaw angle of an object are randomly sampled from a region of $[-4,4]\times[-1,1]\times[20,40]$ and a range of $[-\pi,\pi]$, respectively, and used to calculate the ground truth rotation matrix $\mathbf{R}_{gt}$ and translation vector $\mathbf{t}_{gt}$.
We extract 300 object points and generate 2D image coordinates by projecting them onto the image plane.
Then Gaussian noise of $\sigma=2$~pixels is added to the image coordinates.
The pitch angle of a camera and a ratio of outliers are set to $0^\circ$ and 50\%, respectively.
We design 4 types of experiments.
(1) \textit{E1}: the outlier ratio is changed from 10\% to 90\%.
(2) \textit{E2}: the number of object points is changed from 50 to 1000.
(3) \textit{E3}: pitch angle errors from $-5^\circ$ to $5^\circ$ are added to the camera pose.
(4) \textit{E4}: 2D bounding box errors from -5 pixels to 5 pixels are added to 2D bounding box observations.

\vspace{-9pt}

% Evaluation metric
\subsubsection{Evaluation metric}
\vspace{-6pt}
We use mean rotation and translation errors which have been widely used in the literature~\cite{ferraz2014very,zheng2013revisiting,zheng2013aspnp}.
Given a rotation estimate $\mathbf{R}$ and a translation estimate $\mathbf{t}$, the translation error is measured by $e_{t}(\%)=\| \mathbf{t}_{gt} - \mathbf{t} \| / \| \mathbf{t} \| \times 100$ and the rotation error by $e_{r}(^\circ) = \max^3_{i=1} \{ \arccos{(\mathbf{r}_{gt,i} \cdot \mathbf{r}_i)} \times 180 / \pi \}$ where $\mathbf{r}_{gt,i}$ and $\mathbf{r}_i$ are the $i^{th}$ column of the rotation matrices $\mathbf{R}_{gt}$ and $\mathbf{R}$.
For each experiment, we calculated mean errors of 1000 independent simulations.
The methods were tested on a 3.4GHz single core using MATLAB.

% Figure: Synthetic Exp - Outlier, #inliers, Pitch, Bbox
\begin{figure*}[h]
	\begin{center}
	\includegraphics[width=0.85\linewidth]{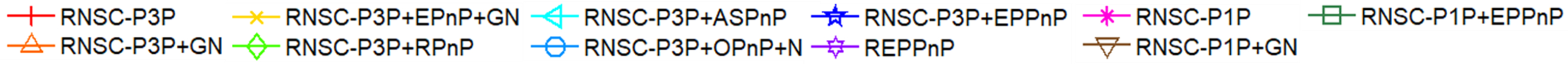} \\ \vspace{-8pt}
	\subfloat[]{
        \includegraphics[width=.29\linewidth,height=100px]{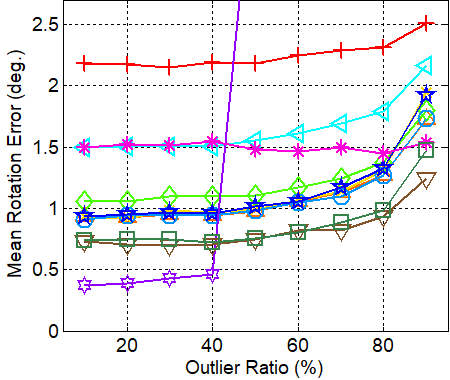}
        \label{fig:exp1_are}
	}
	\subfloat[]{
        \includegraphics[width=.29\linewidth,height=100px]{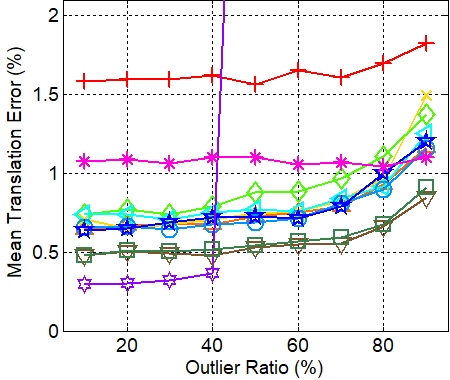}
        \label{fig:exp1_ate}
	}
	\subfloat[]{
        \includegraphics[width=.29\linewidth,height=100px]{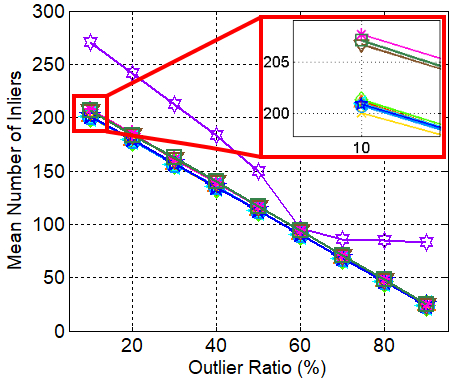}
        \label{fig:exp1_noin}
	} \\ \vspace{-8pt}
	\subfloat[]{
        \includegraphics[width=.29\linewidth,height=100px]{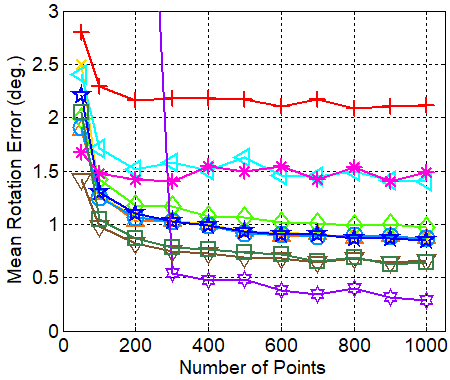}
        \label{fig:exp2_are}
	}
	\subfloat[]{
        \includegraphics[width=.29\linewidth,height=100px]{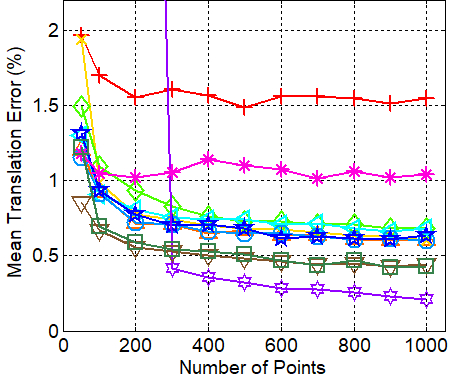}
        \label{fig:exp2_ate}
	}
	\subfloat[]{
        \includegraphics[width=.29\linewidth,height=100px]{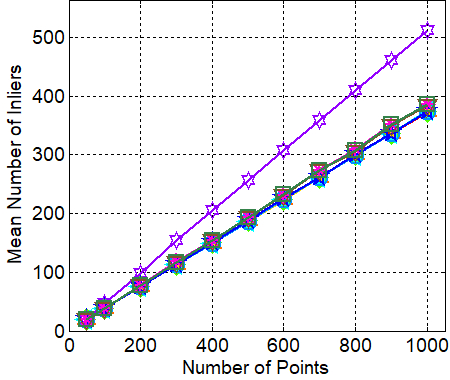}
        \label{fig:exp2_noin}
	} \\ \vspace{-8pt}
	\subfloat[]{
        \includegraphics[width=.21\linewidth,height=100px]{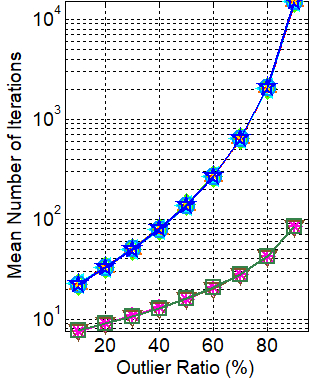}
        \label{fig:exp1_noit}
	}
	\subfloat[]{
        \includegraphics[width=.21\linewidth,height=100px]{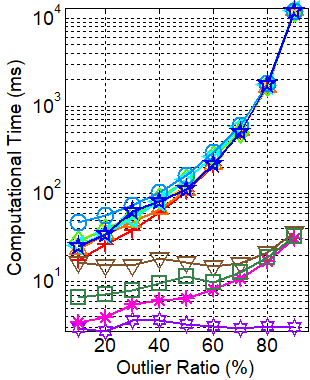}
        \label{fig:exp1_time}
	}
	\subfloat[]{
        \includegraphics[width=.21\linewidth,height=100px]{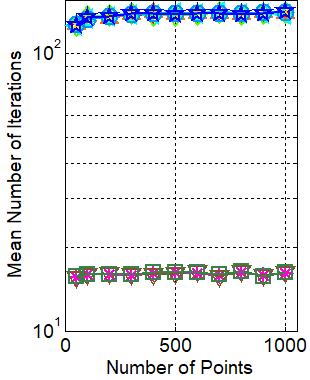}
        \label{fig:exp2_noit}
	}
	\subfloat[]{
        \includegraphics[width=.21\linewidth,height=100px]{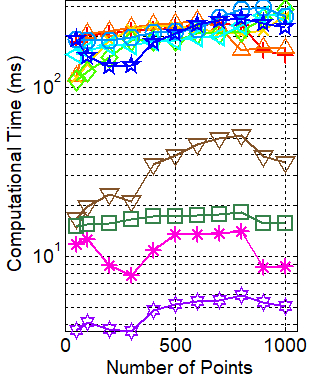}
        \label{fig:exp2_time}
	} \\ \vspace{-8pt}
	\subfloat[]{
        \includegraphics[width=.21\linewidth,height=100px]{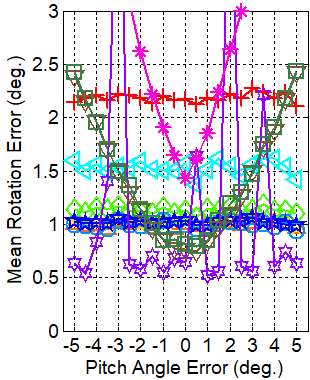}
        \label{fig:exp3_are}
	}
	\subfloat[]{
        \includegraphics[width=.21\linewidth,height=100px]{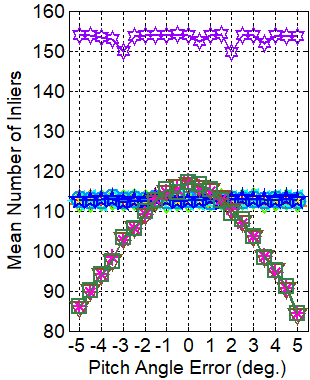}
        \label{fig:exp3_noin}
	}
	\subfloat[]{
        \includegraphics[width=.21\linewidth,height=100px]{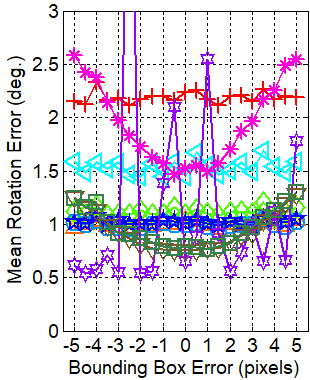}
        \label{fig:exp4_are}
	}
	\subfloat[]{
        \includegraphics[width=.21\linewidth,height=100px]{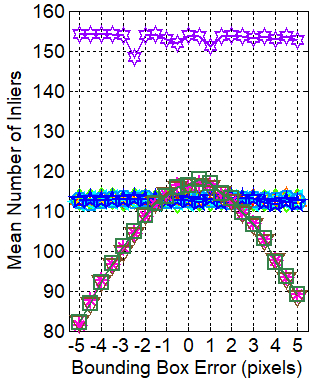}
        \label{fig:exp4_noin}
	} \\ \vspace{-15pt}
	\end{center}
	\caption{Results of synthetic experiments $\textit{E1}$-$\textit{E4}$. (a), (b), (c), (g), and (h) ((d), (e), (f), (i), and (j)) represent the mean rotation errors, the mean translation errors, the mean number of inliers, the mean number of iteration of RANSAC, and the computational time on \textit{E1} (\textit{E2}), respectively. (k) and (l) ((m) and (n)) represent the mean rotation errors and the number of inliers on \textit{E3} (\textit{E4}), respectively. Please see the supplementary material for high-quality images.} \vspace{-13pt}
	\label{fig:syn_exp_12} 
\end{figure*}

\vspace{-9pt}

% Outlier ratio / Num-points
\subsubsection{Variation of the outlier ratio and the number of points}
\vspace{-9pt}
Our 1-point RANSAC-based method (RNSC-P1P) is compared with RANSAC+P3P~\cite{fischler1981random,kneip2011novel} (RNSC-P3P) and REPP$n$P~\cite{ferraz2014very}.
In addition, their results are polished by the Gauss-Newton (GN) method or several P$n$P approaches: EP$n$P~\cite{lepetit2009epnp}, RP$n$P~\cite{li2012robust}, ASP$n$P \cite{zheng2013aspnp}, OP$n$P~\cite{zheng2013revisiting}, and EPP$n$P~\cite{ferraz2014very}.\footnote{The results of EP$n$P and OP$n$P are refined by GN and the Newton (N) method, respectively.}
In all the experiments, the inlier threshold of RANSAC is set to $t_{in} = 4$ pixels and the algebraic error threshold of REPP$n$P is set to $\delta_\mathrm{max} = kt_{in}/f$ where a constant $k = 1.4$ and the focal length $f = 800$ as recommended in~\cite{ferraz2014very}.

Figures~\ref{fig:exp1_are} and \ref{fig:exp1_ate} show the mean rotation and translation errors in \textit{E1}.
It demonstrates that RNSC-P1P (or RNSC-P1P+P$n$P strategies) is superior to RNSC-P3P (or RNSC-P3P+P$n$P strategies).
Since \textit{E1} has no pitch angle error and P1P uses the 1-DoF rotation parameterization constrained by prior pitch information, the pose estimates of P1P are more accurate than those of P3P in this experiment.
Hence, the number of inliers by P1P is higher than that by P3P as shown in Figure~\ref{fig:exp1_noin}.
Figures~\ref{fig:exp2_are} and \ref{fig:exp2_ate} represent the results in \textit{E2}, which show the same tendency as the results of \textit{E1}.
REPP$n$P achieved better accuracy than our method because it took much more inliers by using the loose threshold.
However, REPP$n$P frequently produced invalid object pose estimates in 1) the case that outlier ratio was more than 50\% and 2) the case that a small number of point correspondences were used, \textit{e.g.}, the number of points is less than 200 as shown in Figures~\ref{fig:exp2_are} and \ref{fig:exp2_ate}, because of the effect of the high outlier ratio of 50\%.
% its breakdown point amounts to 50\% as mentioned in \cite{ferraz2014very} and the algebraic outlier analysis in noisy data requires enough number of input data.
On the other hand, RANSAC-based methods consistently provide valid pose estimates despite the high outlier ratio of 90\%.

As shown in Figures~\ref{fig:exp1_noit} and \ref{fig:exp2_noit}, our RNSC-P1P-based methods take much less iterations than RNSC-P3P-based methods.
Consequently, Figures~\ref{fig:exp1_time} and \ref{fig:exp2_time} show that the computational time of RNSC-P1P-based methods is considerably faster than that of RNSC-P3P-based methods but slower than that of REPP$n$P.

\vspace{-9pt}

% Pitch / BBox
\subsubsection{Pitch angle and bounding box errors}
\vspace{-6pt}
Since the proposed method assumed a fixed pitch angle and used a 2D bounding box as additional information, we performed \textit{E3} and \textit{E4} to analyze the effect of the pitch angle and bounding box errors on our method.
Figures~\ref{fig:exp3_are}-\ref{fig:exp4_noin} show that both the pose estimation accuracy and the number of inliers decrease as the pitch angle and bounding box errors increase.
The RNSC-P1P-based methods produce better performance than the RNSC-P3P-based methods if the pitch error is within 1.5 degrees or the bounding box error is within 2 pixel, otherwise their performance is degraded.

% Figure: Synthetic Exp - HRE
\begin{figure*}[t]
	\begin{center}
	\subfloat[]{
        \includegraphics[width=.22\linewidth,height=75px]{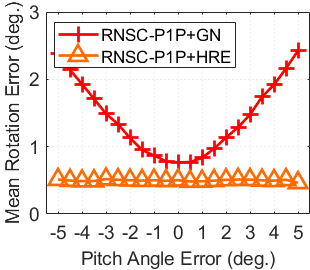}
        \label{fig:exp5_are}
	}
	\subfloat[]{
        \includegraphics[width=.22\linewidth,height=75px]{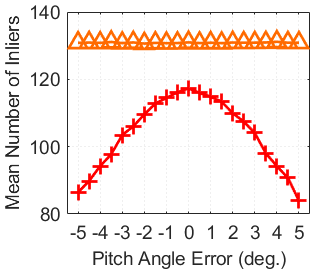}
        \label{fig:exp5_noin}
	}
	\subfloat[]{
        \includegraphics[width=.22\linewidth,height=75px]{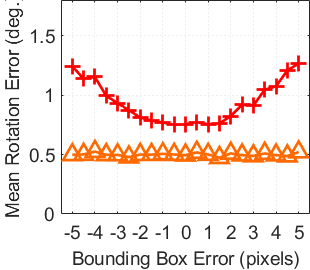}
        \label{fig:exp6_are}
	}
	\subfloat[]{
        \includegraphics[width=.22\linewidth,height=75px]{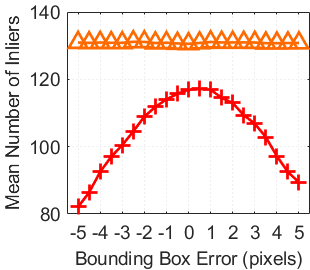}
        \label{fig:exp6_noin}
	} \\ \vspace{-18pt} 
	\end{center}
	\caption{Results of our hierarchical robust estimation method. (a) and (b) ((c) and (d)) represent the mean rotation errors and the number of inliers on \textit{E3} (\textit{E4}), respectively.} \vspace{-12pt}
	\label{fig:syn_exp_hre} 
\end{figure*}

\vspace{-11pt}

% HRE
\subsubsection{Hierarchical robust estimation}
\vspace{-6pt}
As mentioned in Section~\ref{sec:hre}, the hierarchical robust estimation (HRE) method can be used in pose estimation using known 3D models.
We compared RNSC-P1P+HRE with RNSC-P1P+GN in \textit{E3} and \textit{E4}.
In the experiments, the parameters $\tau_1$, $\tau_2$, and $\tau_3$ are set to 4, 6, and 12 pixels, respectively.
As shown in Figures~\ref{fig:exp5_noin} and \ref{fig:exp6_noin}, RNSC-P1P+HRE produces the consistently higher number of inliers than RNSC-P1P+GN in spite of the effect of pitch angle and bounding box errors.
It demonstrates that the scale estimates of HRE are appropriately bounded by manually determined but geometrically meaningful thresholds at each stage.
Consequently, HRE converges to a global minimum even if an initial pose estimate is noisy, whereas the existing estimator converges to local minima, as shown in Figures~\ref{fig:exp5_are} and \ref{fig:exp6_are}.

% GT-INLIER+GN / RNSC-P1P+RE experiments are required more
\vspace{-3pt}

% Virtual driving simulation Dataset
\subsection{Experiments on a virtual simulation dataset} \vspace{-3pt}
\label{sec:exp_virtual}

% Dataset
\subsubsection{Dataset}

\vspace{-6pt}

To evaluate our method in a real application, we generated a dataset from a virtual driving simulation platform.
We defined 53 keypoints for 4 types of vehicles (\ie, car, bus, pickup truck, and box truck) in a similar manner with \cite{song2019apollo}.
Then we generated 2D corresponding keypoints by projecting them using ground truth poses.
Ground truth bounding boxes of objects were calculated from 2D projection of the 3D vehicle models.
The pitch angle $\theta_p$ of the camera was set to $0^\circ$ as prior information.
However, since the camera of the ego-vehicle was considerably shaking and the road surface was often slanted, the pitch angle was regarded to be very noisy.
Images were captured with image resolution of $1914\times1080$ pixels and a horizontal FOV of $50.8^\circ$.
In total, we captured 100,000 frames including 207,977 objects which were split into 142,301, 10,997, and 54,679 instances for training, validation, and testing, respectively.

\vspace{-9pt}

% Keypoint detection
\subsubsection{Keypoint detection} 

\vspace{-6pt}

To detect objects' keypoints, we employ a network architecture similar to \cite{xiao2012simple}. Specifically, we train ResNets\cite{he2016resnet} on top of which three to four transposed Conv-BN-ReLU blocks are stacked to upscale output feature maps. These networks take a cropped object image as input and have two output heads: one outputs prediction maps whose number of channels equals to the maximum number of pre-defined keypoints of all classes; the other predicts the label of the object class of the input. 
To investigate the speed and accuracy trade-offs on various backbones and input resolutions, we used two ResNets (Res50 and Res10\footnote{Res10 models simply remove a residual block in each stage (conv2$\sim$conv5) of Res18.}) as backbones and two resolutions ($256\times 192$ and $128\times 96$) as input sizes.
%Since many classes have common keypoints, the network predicts a single prediction map and the channel index of the same keypoints is consistent across different classes.
%This simple network architecture performs well to generate candidate points as input to the proposed 1-point RANSAC-based method. 
% Applying larger backbones and tailored for keypoint detection
More complicated network architectures~\cite{cheng2019bottom,sun2019deeppose} can boost the detection performance further, but is out of scope of this paper.
%
% We trained and evaluated several keypoint detection networks to investigate the speed and accuracy trade-offs, with various backbones and input resolutions. Specifically, we used three ResNets (Res50, Res18 and Res10\footnote{Res10 models simply remove a residual block in each stage (conv2$\sim$conv5) of Res18.}) as backbones and two resolutions ($256\times 192$ and $128\times 96$) as input sizes.
%Res50 models used three transposed Conv layers to produce prediction maps whose size is 1/4 of the input, all other models predict 1/2 scale prediction maps using four transposed Conv layers. 
To compare the speed of each model, we measured FLOPS and inference speed of models on Qualcomm\textsuperscript{\textregistered} Snapdragon\textsuperscript{TM}\footnote{Qualcomm\textsuperscript{\textregistered} Snapdragon\textsuperscript{TM} is a product of Qualcomm Technologies, Inc. and/or its subsidiaries.} SA8155P's DSP unit. All models were trained by using Adam optimizer~\cite{adam2014adam} with 90 epochs and the learning rate was divided by 10 at 60 and 80 epochs with the initial learning rate of 0.001 and weight decay was set to 0.0001. We applied the softargmax to determine the location of each keypoint. Table~\ref{table:keypoint} shows the comparison of the accuracy and speed among these models. 

\setlength{\tabcolsep}{0.8ex}
{
\begin{table}[t]
\caption{Speed and accuracy trade-offs of various keypoint detection networks.} \vspace{-15pt}
\begin{center}
\small
\begin{tabular}{ccccc}
\toprule
\multirow{2}*{Model} & \multirow{2}*{Class acc.} & Vertex err. &  \multirow{2}*{GMAC} & Latency \\
& & (pixel) & & (ms)\\ \midrule
\multicolumn{1}{l}{Res50 $256\times192$} & 0.997 & 5.24 & 5.45 & 13.54 \\
%\multicolumn{1}{l}{Res18 $256\times192$} & 0.995 & 6.41 & 3.23 & 7.43 \\
\multicolumn{1}{l}{Res10 $256\times192$} & 0.990 & 7.12 & 2.32 & 5.84 \\
\multicolumn{1}{l}{Res50 $128\times96$} & 0.993 & 7.57 & 1.36 & 7.31 \\
%\multicolumn{1}{l}{Res18 $128\times96$} & 0.991 & 9.36 & 0.81 & 4.22 \\
\multicolumn{1}{l}{Res10 $128\times96$} & 0.986 & 10.05 & 0.50 & 3.03 \\ \toprule
%Res10 $128\times96$ comp. & 0.982 & 11.11 & 0.42 & 2.97 \\
\end{tabular} \vspace{-15pt}
\end{center}
\label{table:keypoint}
\end{table}
}

% Table
{
\setlength{\tabcolsep}{0.6ex}
\begin{table}[t]
\caption{Rotation, translation, and vertex errors for the experiments on known models (\textit{G1}) and ASM (\textit{G2}). \textbf{Bold} and \textit{italic} mean the best and the second best, respectively.} \vspace{-15pt}
\begin{center}
\footnotesize
\begin{tabular}{l l cc cc l cc cc}
\toprule
\multirow{3}*{Model} & & \multicolumn{4}{l}{Known models (\textit{G1})} & & \multicolumn{4}{l}{ASM (\textit{G2})} \\ \cmidrule(r){3-6} \cmidrule{8-11}
 & & \multicolumn{2}{l}{RNSC-P1P} &\multicolumn{2}{l}{RNSC-P3P} & &  \multicolumn{2}{l}{RNSC-P1P} &\multicolumn{2}{l}{RNSC-P3P} \\ \cmidrule(r){3-4} \cmidrule(r){5-6} \cmidrule(r){8-9} \cmidrule{10-11}
 & & \multicolumn{1}{l}{GN} & \multicolumn{1}{l}{HRE} & \multicolumn{1}{l}{GN} & \multicolumn{1}{l}{HRE} & & \multicolumn{1}{l}{RE} & \multicolumn{1}{l}{HRE} & \multicolumn{1}{l}{RE} & \multicolumn{1}{l}{HRE} \\
\midrule
\multirow{2}*{Res10} & $e_r$ & 3.78 & \textit{1.56} & 1.90 & \textbf{1.41} & $e_r$ & 2.98  & \textbf{2.20} & 3.68 & \textit{2.77} \\
& $e_t$ & 6.28 & \textit{1.61} & 1.76 & \textbf{1.30} & $e_v$ & 30.53 & \textbf{21.16} & 23.08 & \textit{21.28} \\
\multirow{2}*{Res50} & $e_r$ & 3.76 & \textit{1.19} & 1.46 & \textbf{0.96} & $e_r$ & 2.57 & \textbf{1.96} & 3.16 & \textit{2.43} \\
& $e_t$ & 6.38 & \textit{1.36} & 1.41 & \textbf{0.93} & $e_v$ & 21.43 & \textit{21.23} & 22.18 & \textbf{20.77} \\\toprule
\end{tabular} \vspace{-20pt}
\end{center}
\label{table:simulation_eval}
\end{table}
}

\vspace{-9pt}

% Evaluation
\subsubsection{Evaluation}

\vspace{-6pt}

We performed two types of experiments: (1) \textit{G1}: object pose estimation using known 3D object models and (2) \textit{G2}: object pose and shape estimation using ASM.
In \textit{G1}, we evaluate RNSC-P1P and RNSC-P3P with GN and HRE, respectively, and measure the average rotation and translation errors.
In \textit{G2}, we take the same protocol with \textit{G1} but substitute GN with the robust estimation (RE) method of Eq.~(\ref{eq:re}) and additionally measure average vertex error $e_v(\mathrm{cm})$ between ground truth and reconstructed 3D models whose scales are adjusted using a scale difference between ground truth and estimated translation vectors because of its scale ambiguity.
In the experiments, the ground truth 2D bounding boxes were used and the parameters $t_{in}$, $\tau_1$, $\tau_2$, and $\tau_3$ were set to $0.0375l$, $0.0375l$, $0.05l$, and $0.15l$, respectively, where $l =\max(w_o,h_o)$ and $w_o$ and $h_o$ were width and height of a 2D object bounding box.
Table~\ref{table:simulation_eval} presents the results for the experiments.
It shows that the performance of HRE is superior to those of both GN and RE.
In \textit{G1}, RNSC-P3P-based methods are more accurate than RNSC-P1P-based methods due to pitch errors.
Nevertheless, the performance of RNSC-P1P+HRE surpasses that of RNSC-P3P+GN by virtue of securing more valid inliers by HRE.
\textit{G2} is a more difficult scenario because the pose should be calculated from an inaccurate 3D model, \textit{i.e.}, the mean shape of ASM.
Contrary to \textit{G1}, RNSC-P1P-based methods achieve better rotational accuracy as the constrained pitch angle rather restricts the range of a rotation estimate in the early optimization stage with large shape variation.

\vspace{-9pt}

% Computational time
\subsubsection{Computational time}

\vspace{-6pt}

We tested the proposed method, \textit{i.e.}, RNSC-P1P+HRE with the keypoint detection model of Res10-$128\times96$ on the DSP unit. Given a 2D object bounding box, the pose and shape estimation took 4.38~ms per object where RNSC-P1P+HRE took 0.15~ms on CPU, keypoint detection took 3.03~ms on DSP, and the pre- and post-processing, such as normalization, image cropping and resizing, and softargmax operation for keypoint extraction, took the rest of the computational time. 

\vspace{-6pt}

% Real-world Dataset
\subsection{Experiments on real-world datasets} \vspace{-6pt}

% Experimental setting
\subsubsection{Experimental setting}
\vspace{-6pt}
We used the KITTI object detection dataset~\cite{geiger2012arewe} to evaluate quantitatively our method in real-world scenes.
Following the protocol of \cite{xiaozhi2017multiview}, we split the KITTI training data into the \textit{training} and \textit{validation} sets.
From the training set, we selected 764 and 1019 instance samples for training and validation of the keypoint detection network, respectively. We manually annotated keypoints and then trained the model of Res50~$256\times192$.
In addition, we captured images in a real-world scene to qualitatively compare the methods.

\vspace{-6pt}

% Table
\begin{table}[t]
\caption{Rotation and translation errors for the experiments on the KITTI validation set. Each value represents an error at the easy/moderate/hard case.} \vspace{-15pt}
\begin{center}
\small
\begin{tabular}{l c@{}c@{}c@{}c@{}c c@{}c@{}c@{}c@{}c }
\toprule
\multicolumn{1}{c}{Method} & \multicolumn{5}{c}{$e_r(^\circ)$} & \multicolumn{5}{c}{$e_a(^\circ)$} \\ \midrule
RNSC-P1P+RE & \textbf{3.203} &/& \textit{4.141} &/& \textit{4.283} & \textit{0.1376} &/& \textit{0.1509} &/& \textit{0.1531} \\
RNSC-P1P+HRE & \textit{3.365} &/& \textbf{4.016} &/& \textbf{4.071} & \textbf{0.1356} &/& \textbf{0.1450} &/& \textbf{0.1460} \\
RNSC-P3P+RE & 4.083 &/& 5.482 &/& 5.450 & 0.1511 &/& 0.1651 &/& 0.1644 \\
RNSC-P3P+HRE & 4.021 &/& 5.243 &/& 5.231 & 0.1472 &/& 0.1616 &/& 0.1613 \\ \toprule
\end{tabular} \vspace{-15pt}
\end{center}
\label{table:real_eval}
\end{table}

% Evaluation
\subsubsection{Evaluation}
\begin{figure}[t]
	\begin{center}
	\subfloat[]{
        \includegraphics[width=.48\linewidth,height=110px]{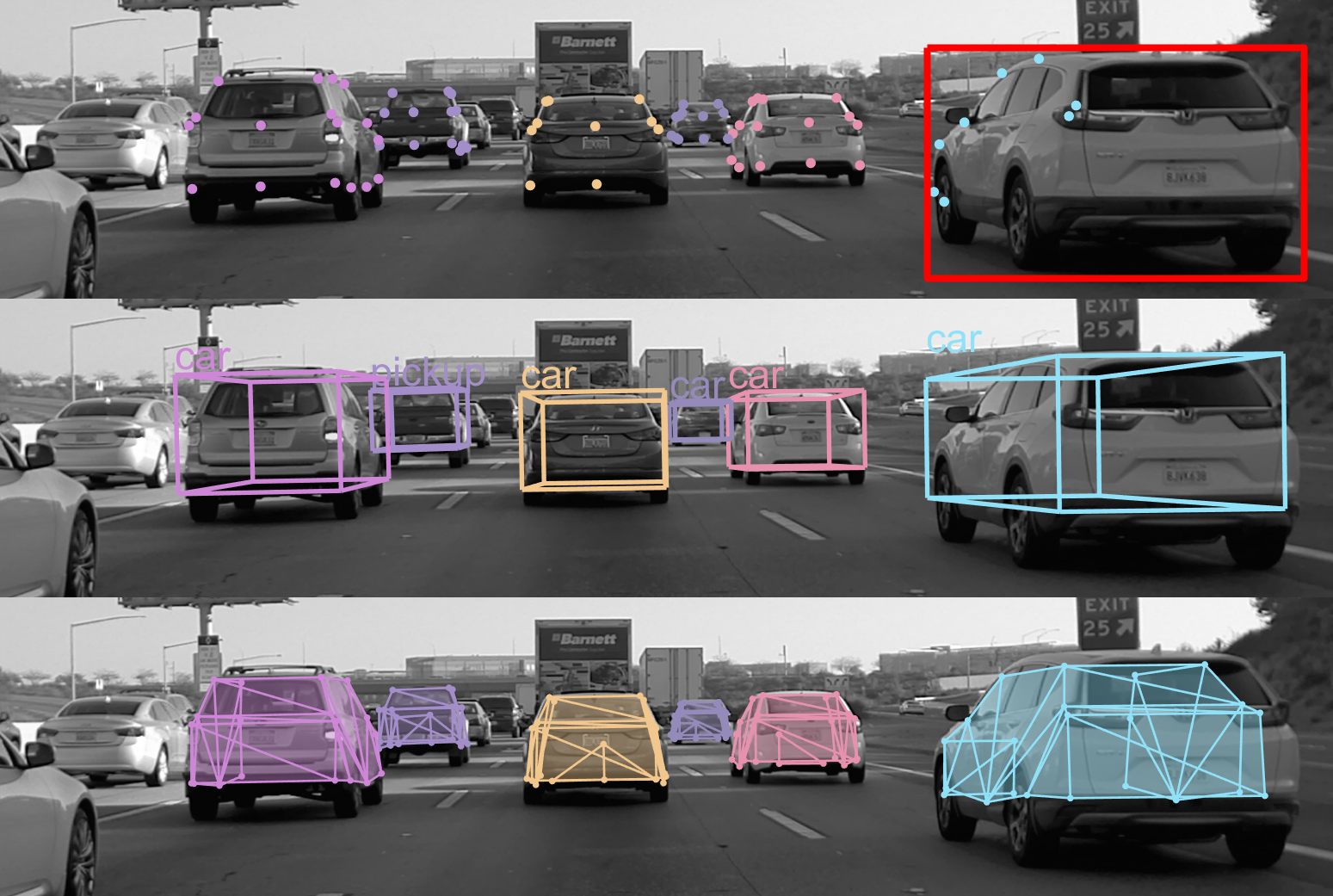}
        \label{fig:simulation_re}
	}
	\subfloat[]{
        \includegraphics[width=.48\linewidth,height=110px]{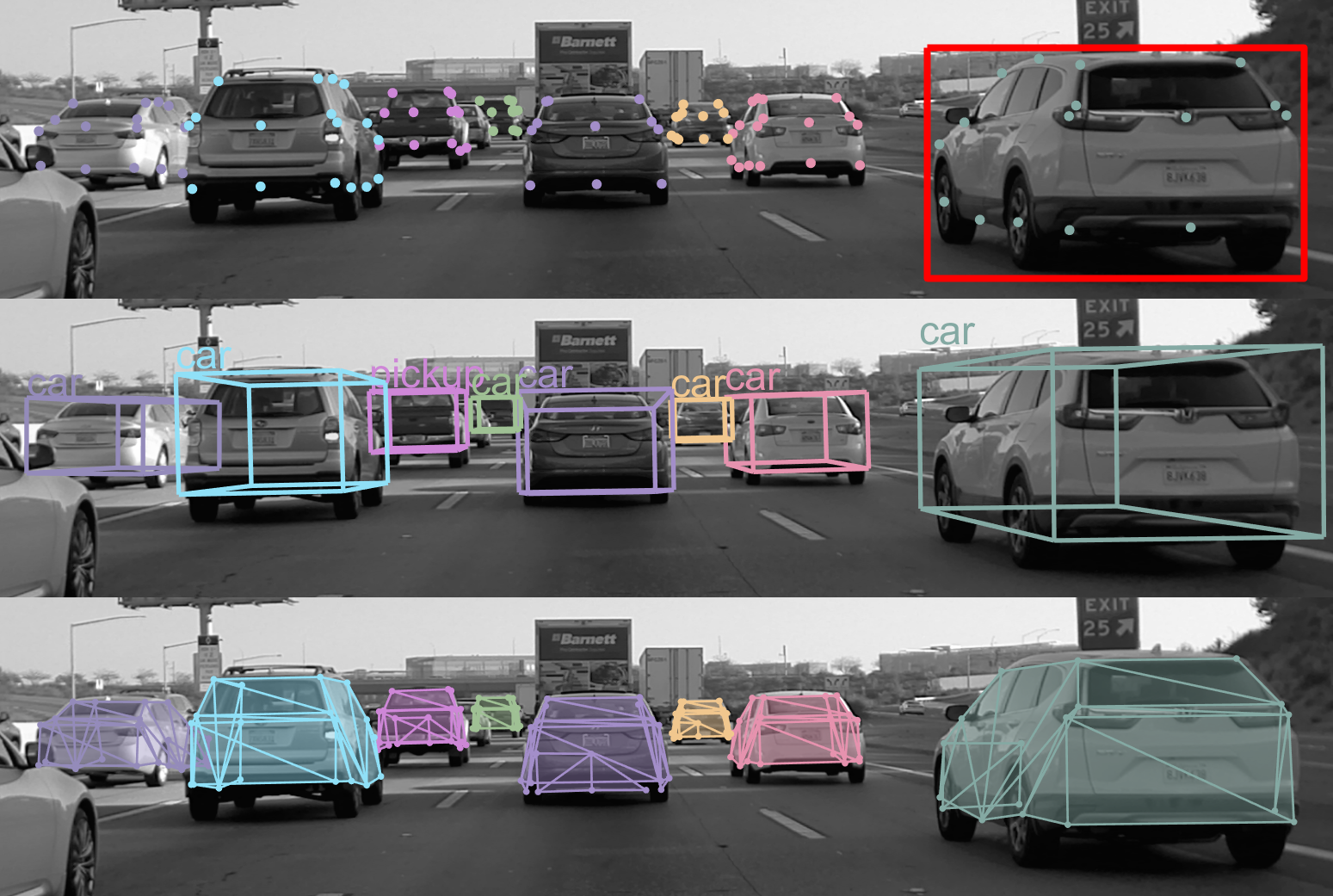}
        \label{fig:simulation_hre}
	} \\ \vspace{-15pt} 
	\end{center}
	\caption{Results of RNSC-P1P+RE(a) and RNSC-P1P+HRE(b). Top, middle, and bottom images represent 2D projection of inlier keypoints, 3D bounding boxes, and shape reconstruction results using ASM, respectively.} \vspace{-8pt}
	\label{fig:simulation_res} 
\end{figure}

\vspace{-6pt}
We evaluate RNSC-P1P and RNSC-P3P with RE and HRE, respectively, on the validation set by measuring rotation and translation errors as in Section~\ref{sec:exp_virtual}. 
However, since the translation estimate has scale ambiguity, we employ the average angular error between $\mathbf{t}$ and $\mathbf{t}_{gt}$ as a translation error $e_a(^\circ)$ according to \cite{jiaolong2014optimal}.
As shown in Table~\ref{table:real_eval}, RNSC-P1P+HRE achieves the best performance in most cases.
Figure~\ref{fig:simulation_res} shows the results of RNSC-P1P+RE and RNSC-P1P+HRE from an input image captured in a real-world scene.
In the red box of Figure~\ref{fig:simulation_re}, RNSC-P1P+RE reconstructed the shape of the vehicle incorrectly, whereas RNSC-P1P+HRE estimated its shape completely with more inlier keypoints.
It demonstrates that the proposed method works well in practical applications that require to detect objects with arbitrary shape under abrupt pitch angle variation by a shaking camera.

% Figure: Synthetic Exp - HRE

\vspace{-6pt}

% ============== Conclusion
\section{Conclusion} \vspace{-6pt}

In this paper, we proposed an efficient method based on 1-point RANSAC to estimate a pose of an object on the ground.
Our 1-point RANSAC-based method using 2D bounding box prior information was much faster than the conventional method such as RANSAC+P3P by significantly reducing the number of trials.
In addition, our hierarchical robust estimation method using geometrically meaningful and multiple scale estimates produced superior results in the evaluation using synthetic, virtual driving simulation, and real-world datasets.

{\small
\bibliographystyle{ieee_fullname}
\bibliography{egbib}
}

% ============== Appendix
\newpage
\onecolumn
\appendix

% ============== Appendix A: RANSAC
\section{RANSAC-based scheme for object pose estimation}

\setcounter{algorithm}{0}
\renewcommand{\thealgorithm}{A.\arabic{algorithm}}

A general framework for $n$-point RANSAC-based pose estimation is shown in Alg.~\ref{alg:ransac}.

\begin{algorithm} [h]
	\caption{Pose estimation using RANSAC and a P$n$P algorithm}
	\label{alg:ransac}
	\begin{algorithmic}[1]
		\REQUIRE $\mathcal{C}$
		\ENSURE $\mathbf{T}$
		\STATE $n_{best} \leftarrow 0, \mathbf{T} \leftarrow \varnothing$
		\FOR{$t = 1$ \TO $N$}
		\STATE Randomly sample $n$ 2D-3D keypoint correspondences from $\mathcal{C}$.
		\STATE $\mathbf{T}_{cand}$ $\leftarrow$ Compute a pose candidate using the P$n$P algorithm from the $n$ samples.
		\STATE $n_{cand} \leftarrow$ Count the number of inliers using $\mathbf{T}_{cand}$.
		\IF{$n_{cand} > n_{best}$}
		\STATE $\mathbf{T} \leftarrow \mathbf{T}_{cand}, n_{best} = n_{cand}$
		\ENDIF
		\ENDFOR
		\IF{$\mathbf{T} \neq \varnothing$}
		\STATE $\mathbf{T} \leftarrow$ Refine $\mathbf{T}$ using the inlier points.
		\ENDIF
	\end{algorithmic}
\end{algorithm}

% ============== Appendix B: P1P derivation
\section{Details of perspective-1-point solution}

\setcounter{figure}{0}
\setcounter{equation}{0}
\setcounter{table}{0}
\renewcommand{\theequation}{B.\arabic{equation}}
\renewcommand{\thetable}{B.\arabic{table}}
\renewcommand{\thefigure}{B.\arabic{figure}}

% Definition of pitch angle
\subsection{Definition of pitch angle}

\begin{figure}[h]
	\begin{center}
		\includegraphics[width=.28\linewidth]{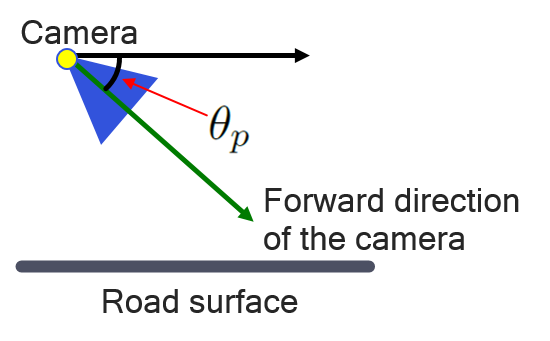} \vspace{-10pt}
	\end{center}
	\caption{Definition of pitch angle} \vspace{-3pt}
	\label{fig:def_pitch} 
\end{figure}

Since we want to simplify the P$n$P problem, we redefine the problem in the bird-eye view (BEV) in this paper.
Figure~\ref{fig:def_pitch} shows the definition of a pitch angle $\theta_p$ of a camera.
Once the pitch angle between the camera and the road surface is known, a 3D directional vector $\mathbf{V}_i$, \textit{i.e.}, a back-projected ray of a pixel on an image, is approximated to a 2D directional vector $\mathbf{v}_i$ on the top-view as follows.

\begin{equation}
    \label{eq:2dvec}
    \mathbf{v}_i = \begin{bmatrix} V_X \\ V_Z \end{bmatrix}, \ \mathrm{where} \ \begin{bmatrix} V_X \\ V_Y \\ V_Z \end{bmatrix} = \mathbf{R}^{-1}_{cg} \mathbf{V}_i
\end{equation}

% Parameter calculation
\subsection{Parameter calculation}

Many parameters\footnote{Please see the details of the parameters in Fig.~3.} such as $\phi_y$, $\theta_L$, $\theta_R$, $\psi_L$, $\psi_R$, $l_L$, and $l_R$ should be calculated prior to derivation of Eqs.~(1)~and~(2).
The parameters $\phi_y$, $\theta_L$, and $\theta_R$ are computed as follows.

\begin{eqnarray}
\phi_y = \arcsin{\mathbf{v}_C \cdot \mathbf{v}_K} \\
\theta_L = \arccos{\mathbf{v}_K \cdot \mathbf{v}_L} \\
\theta_R = \arccos{\mathbf{v}_K \cdot \mathbf{v}_R}
\end{eqnarray}

Let's denote the $i^{th}$ corner point as $\mathbf{p}_i$ and a selected keypoint as $\mathbf{p}_K$. Then, the parameters $\psi_L$, $\psi_R$, $l_L$, and $l_R$ are calculated depending on a case as the following table.

\setlength{\tabcolsep}{4pt}
{
\renewcommand{\arraystretch}{1.4}
\begin{table}[h] 
    \centering
    \small
    \caption{Parameter computation for the four cases}
    \begin{tabular}{ccccc} \toprule
          & Case 1 & Case 2 & Case 3 & Case 4  \\\midrule
         $l_L$ & $\left \| \mathbf{p}_K - \mathbf{p}_4 \right \|$ & $\left \| \mathbf{p}_K - \mathbf{p}_3 \right \|$ & $\left \| \mathbf{p}_K - \mathbf{p}_1 \right \|$ & $\left \| \mathbf{p}_K - \mathbf{p}_2 \right \|$ \\
         $l_R$ & $\left \| \mathbf{p}_K - \mathbf{p}_2 \right \|$ & $\left \| \mathbf{p}_K - \mathbf{p}_1 \right \|$ & $\left \| \mathbf{p}_K - \mathbf{p}_3 \right \|$ & $\left \| \mathbf{p}_K - \mathbf{p}_4 \right \|$ \\
         $\psi_L$ & $-\mathrm{acos} \ \mathbf{v}_f \cdot \frac{\mathbf{p}_4 - \mathbf{p}_K}{l_L}$ & $\mathrm{acos} \ \mathbf{v}_f \cdot \frac{\mathbf{p}_3 - \mathbf{p}_K}{l_L}$ & $-\mathrm{acos} \ \mathbf{v}_f \cdot \frac{\mathbf{p}_1 - \mathbf{p}_K}{l_L}$ & $\mathrm{acos} \ \mathbf{v}_f \cdot \frac{\mathbf{p}_2 - \mathbf{p}_K}{l_L}$ \\
         $\psi_R$ & $\mathrm{acos} \ \mathbf{v}_f \cdot \frac{\mathbf{p}_2 - \mathbf{p}_K}{l_R}$ & $-\mathrm{acos} \ \mathbf{v}_f \cdot \frac{\mathbf{p}_1 - \mathbf{p}_K}{l_R}$ & $\mathrm{acos} \ \mathbf{v}_f \cdot \frac{\mathbf{p}_3 - \mathbf{p}_K}{l_R}$ & $-\mathrm{acos} \ \mathbf{v}_f \cdot \frac{\mathbf{p}_4 - \mathbf{p}_K}{l_R}$ \\ \toprule
    \end{tabular}
    \label{tab:params}
\end{table} 
}

% Derivation
\subsection{Derivation of P1P solution}

Here, we show the process of the derivation of Eq.~(2).
First of all, we prove Case~1.
From the sine rule,
\begin{equation} \label{eq:sinerule}
    \frac{l_L}{\sin{\theta_L}} = \frac{l_C}{\sin{\phi_L}} \ \mathrm{and} \ \frac{l_R}{\sin{\theta_R}} = \frac{l_C}{\sin{\phi_R}},
\end{equation}
an equation is formulated as follows.
\begin{equation}
    l_C = \frac{l_R \sin{\phi_R}}{\sin{\theta_R}} = \frac{l_L \sin{\phi_L}}{\sin{\theta_L}} \label{eq:sinerule2}
\end{equation}
Because $\phi_R = \theta_y + \psi_R - \theta_R$ ($\phi_R > 0$) and $\phi_L = -\theta_y - \psi_L - \theta_L$ ($\phi_L > 0$), Eq.~(\ref{eq:sinerule2}) is substituted to
\begin{equation}
    \frac{l_R \sin{(\theta_y + \psi_R - \theta_R)}}{\sin{\theta_R}} = \frac{l_L \sin{(-\theta_y - \psi_L - \theta_L)}}{\sin{\theta_L}}. \label{eq:sinerule3}
\end{equation}
Let $\omega_R = \psi_R - \theta_R$ and $\omega_L = \psi_L + \theta_L$. According to the angle addition and subtraction formulae, the sine functions are decomposed as
\begin{equation}
    l_R\frac{\sin{\theta_y}\cos{\omega_R}+\cos{\theta_y}\sin{\omega_R}}{\sin{\theta_R}} = l_L\frac{\sin{\theta_y}\cos{\omega_L} - \cos{\theta_y}\sin{\omega_L}}{\sin{\theta_L}}, \label{eq:formula1}
\end{equation}
because $\psi_R - \theta_R > 0$ and $-\psi_L -\theta_L > 0$. 
Eq.~(\ref{eq:formula1}) is reorganized with respect to $\sin \theta_y$ and $\cos \theta_y$ as 
\begin{equation}
    \sin{\theta_y} \left ( \frac{l_R\cos{\omega_R}}{\sin{\theta_R}} + \frac{l_L\cos{\omega_L}}{\sin{\theta_L}} \right )  = \cos{\theta_y} \left ( -\frac{l_R\sin{\omega_R}}{\sin{\theta_R}} - \frac{l_L\sin{\omega_L}}{\sin{\theta_L}} \right ). \label{eq:formula2}
\end{equation}
Finally, we reorganize Eq.~(\ref{eq:formula2}) with respect of $\theta_y$ as
\begin{equation} \label{eq:yaw}
    \theta_y = \arctan{\left( \frac{-\frac{l_R \sin{\left(\psi_R - \theta_R\right)}}{\sin{\theta_R}} - \frac{l_L \sin{\left(\psi_L + \theta_L\right)}}{\sin{\theta_L}}}{\frac{l_R \cos{\left(\psi_R - \theta_R\right)}}{\sin{\theta_R}} + \frac{l_L \cos{\left(\psi_L + \theta_L\right)}}{\sin{\theta_L}}} \right)}.
\end{equation}

For Cases~2-4, the local yaw angle $\theta_y$ is computed in the same way of Eqs.~(\ref{eq:sinerule})-(\ref{eq:yaw}).

% ============== Appendix C: Keypoint definition
\newpage
\section{Definition of keypoints}

\setcounter{figure}{0}
\renewcommand{\thefigure}{C.\arabic{figure}}

\begin{figure}[hp]
	\begin{center}
	\subfloat[]{
        \includegraphics[width=.8\linewidth,height=120px]{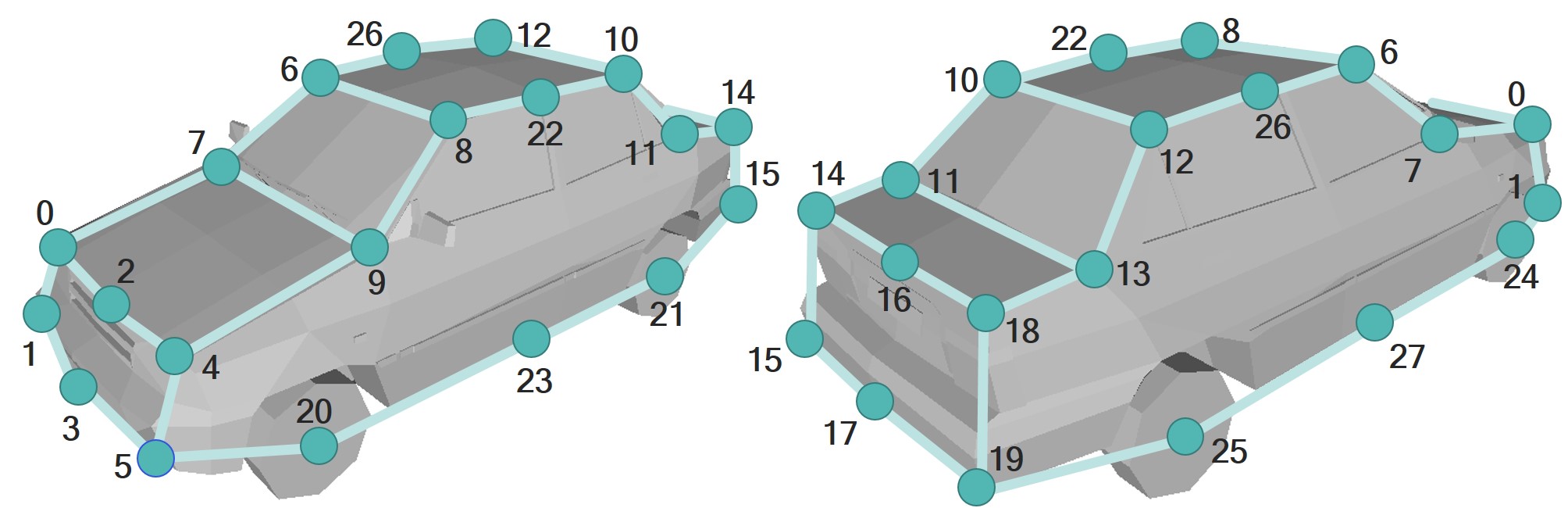} \vspace{-10pt}
        \label{fig:keypt_car} 
	} \\ \vspace{-5pt}
	\subfloat[]{
        \includegraphics[width=.8\linewidth,height=120px]{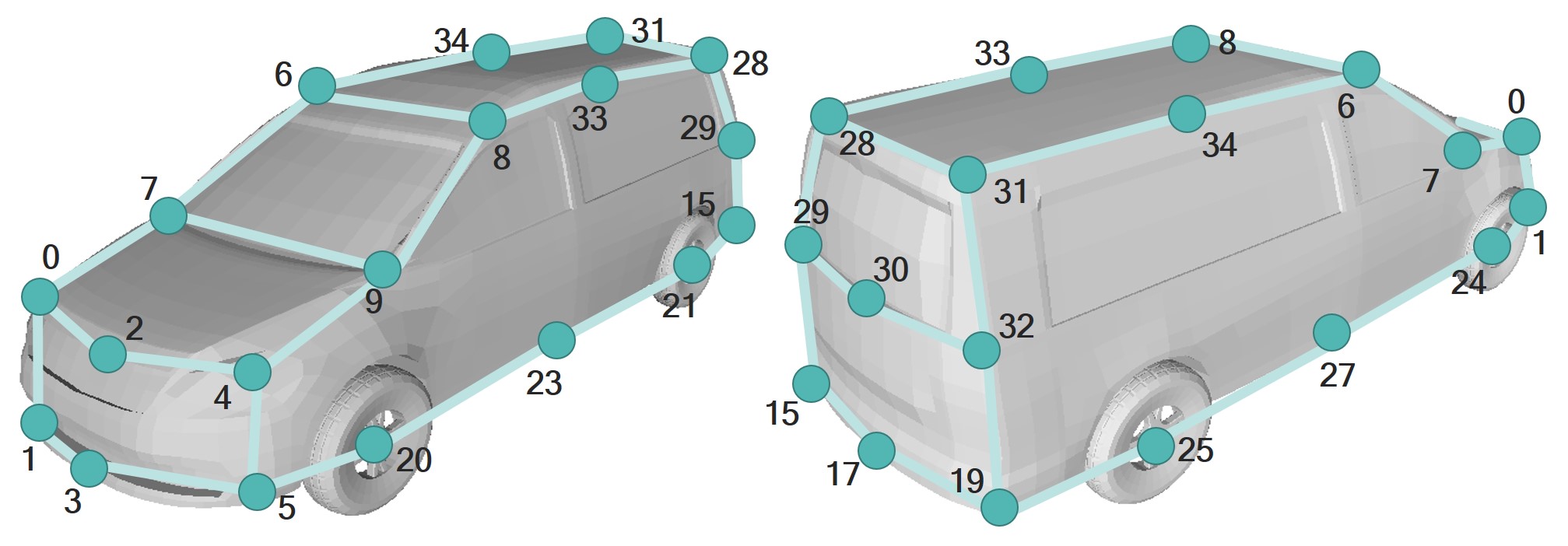} \vspace{-10pt}
        \label{fig:keypt_vanbus}
	} \\ \vspace{-5pt}
	\subfloat[]{
        \includegraphics[width=.8\linewidth,height=120px]{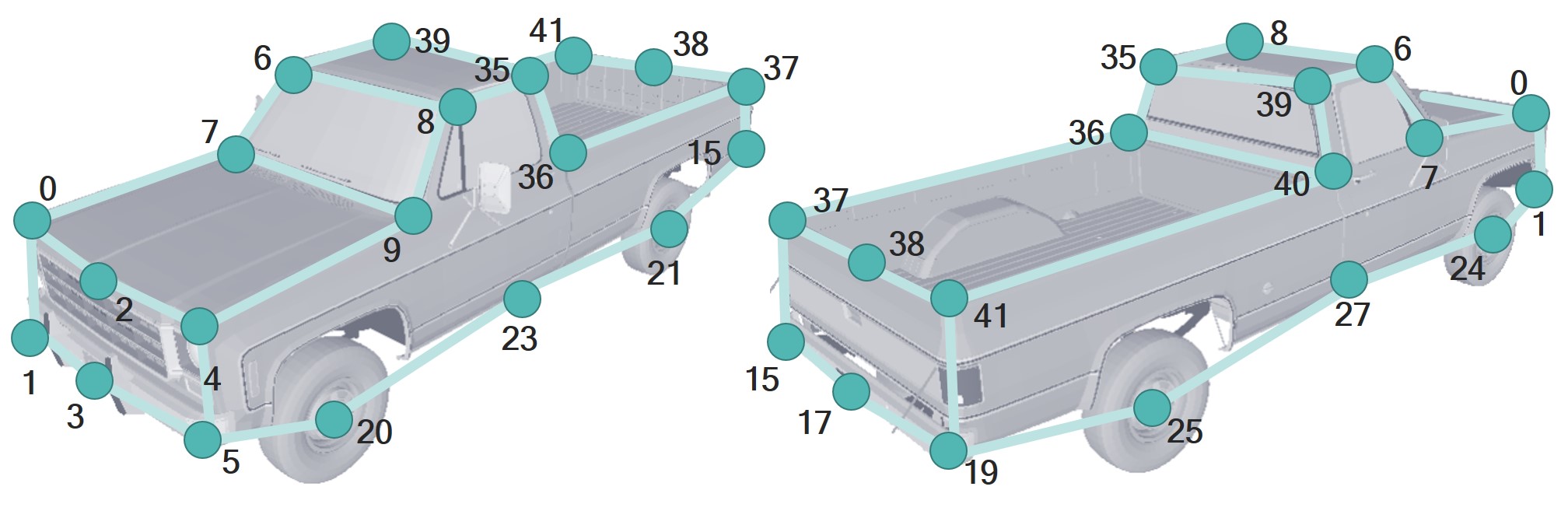} \vspace{-10pt}
        \label{fig:keypt_pick} 
	} \\ \vspace{-5pt}
	\subfloat[]{
        \includegraphics[width=.8\linewidth,height=120px]{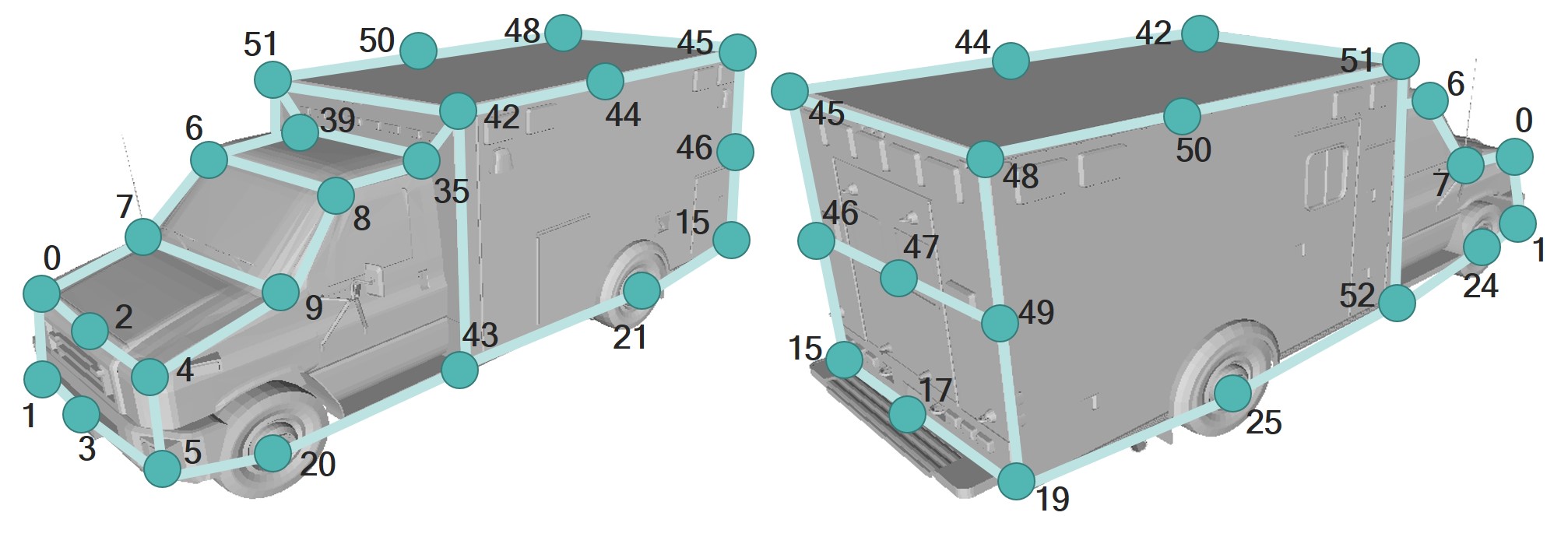} \vspace{-10pt}
        \label{fig:keypt_box} 
	} \\ \vspace{-8pt}
	\end{center}
	\caption{Definition of keypoints of \ccar, \cvanbus, \cpickup, and \cbox} \vspace{-8pt}
	\label{fig:keypt_def} 
\end{figure}

We define 53 keypoints and 4 types of vehicles: \ccar, \cvanbus, \cpickup, and \cbox.
The numbers of keypoints of \ccar, \cvanbus, \cpickup, and \cbox~are 28, 26, 26, and 30, respectively. 
Some keypoints are shared among the classes.
The location and index of each keypoint are shown in Fig.~\ref{fig:keypt_def}.

% ============== Appendix D: Keypoint definition
\newpage
\section{Definition of keypoints}
\vspace{-8pt}
\setcounter{figure}{0}
\renewcommand{\thefigure}{D.\arabic{figure}}
\begin{figure*}[h]
	\begin{center}
	\includegraphics[width=0.85\linewidth]{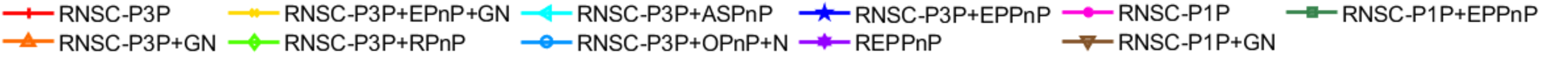} \\ \vspace{-8pt}
	\subfloat[]{
        \includegraphics[width=.45\linewidth]{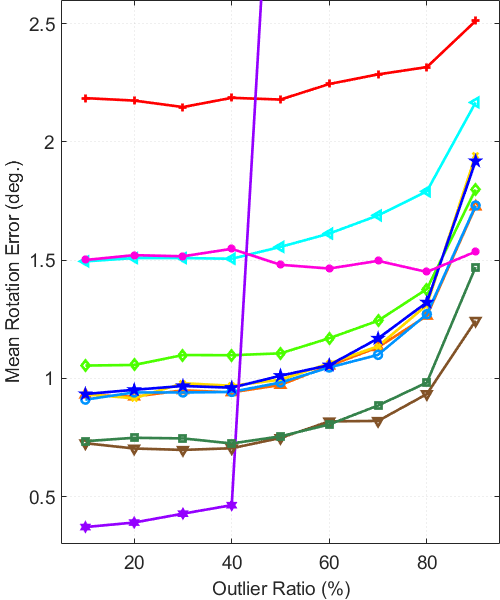}
	}
	\subfloat[]{
        \includegraphics[width=.45\linewidth]{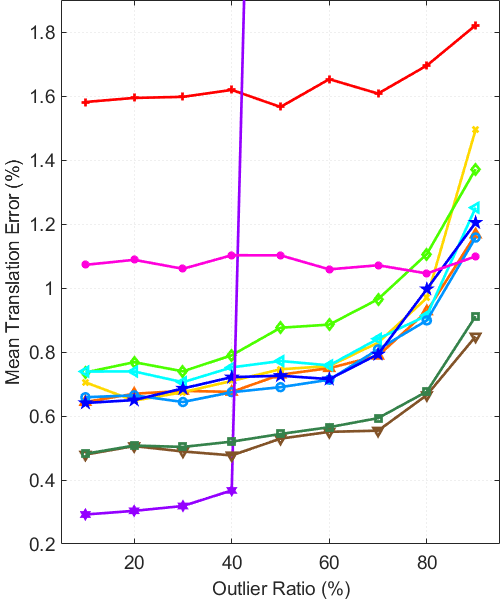}
	} \\ \vspace{-8pt}
		\subfloat[]{
        \includegraphics[width=.45\linewidth]{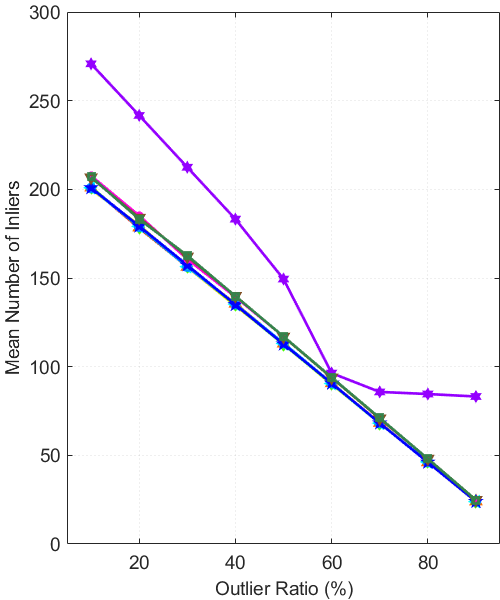}
	}
	\end{center} \vspace{-15pt}
	\caption{High-quality images of Figs. 4(a)-4(c), i.e., the mean rotation errors, the mean translation errors, and the mean number of inliers on \textit{E1}, respectively}
\end{figure*}

\begin{figure*}[h]
	\begin{center}
	\includegraphics[width=0.85\linewidth]{figures/legend_supp.png} \\ \vspace{-8pt}
	\addtocounter{subfigure}{3}
	\subfloat[]{
        \includegraphics[width=.45\linewidth]{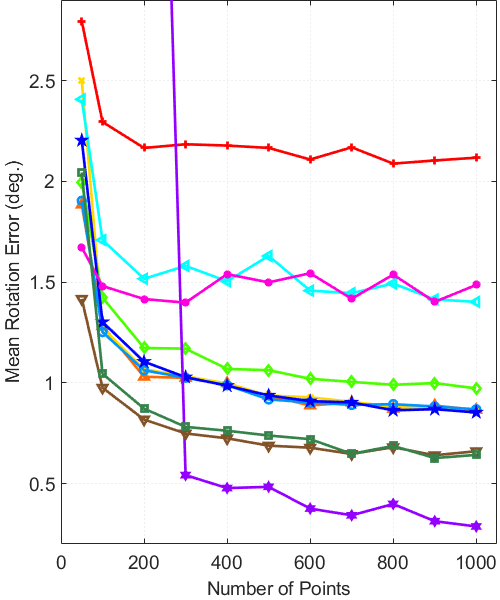}
	}
	\subfloat[]{
        \includegraphics[width=.45\linewidth]{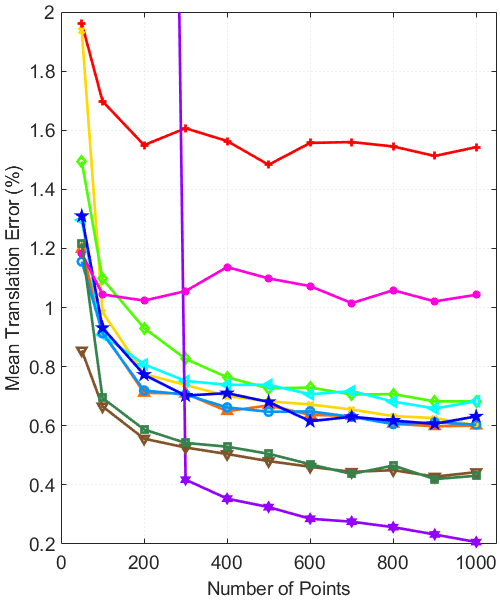}
	} \\
		\subfloat[]{
        \includegraphics[width=.45\linewidth]{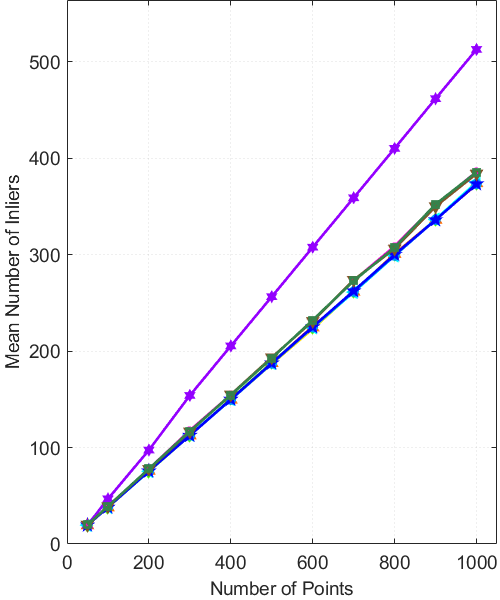}
	}
	\end{center}
	\caption{High-quality images of Figs. 4(d)-4(f), i.e., the mean rotation errors, the mean translation errors, and the mean number of inliers on \textit{E2}, respectively}
\end{figure*}

\begin{figure*}[h]
	\begin{center}
	\includegraphics[width=0.85\linewidth]{figures/legend_supp.png} \\ \vspace{-8pt}
	\addtocounter{subfigure}{6}
	\subfloat[]{
        \includegraphics[width=.45\linewidth]{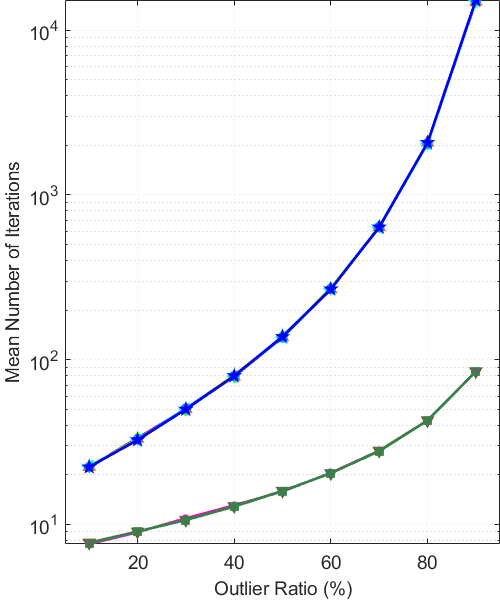}
	}
	\subfloat[]{
        \includegraphics[width=.45\linewidth]{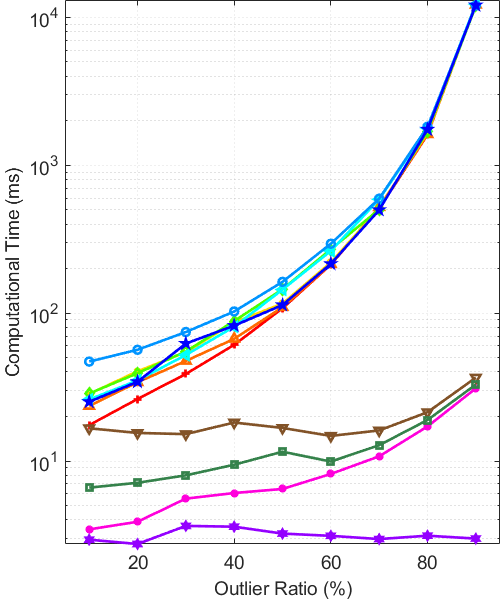}
	} \\
		\subfloat[]{
        \includegraphics[width=.45\linewidth]{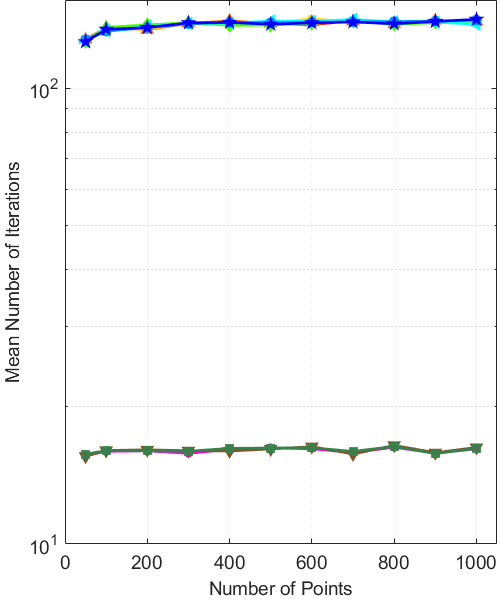}
	}
	\subfloat[]{
        \includegraphics[width=.45\linewidth]{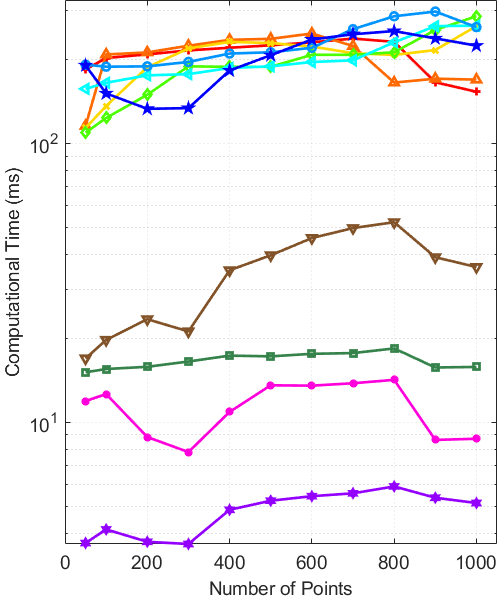}
	} 
	\end{center}
	\caption{High-quality images of Figs. 4(g)-4(j), i.e., the mean number of iteration of RANSAC and the computational time on \textit{E1} and \textit{E2}, respectively}
\end{figure*}

\begin{figure*}[h]
	\begin{center}
	\includegraphics[width=0.85\linewidth]{figures/legend_supp.png} \\ \vspace{-8pt}
	\addtocounter{subfigure}{10}
	\subfloat[]{
        \includegraphics[width=.45\linewidth]{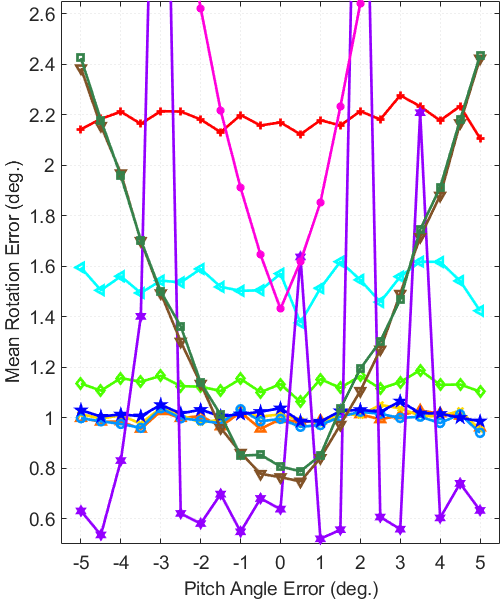}
	}
	\subfloat[]{
        \includegraphics[width=.45\linewidth]{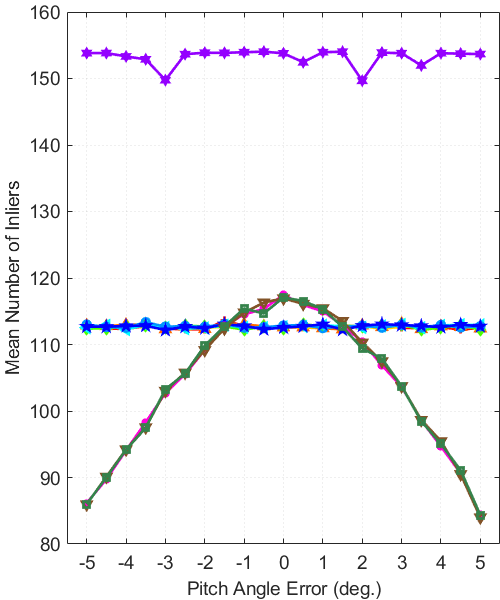}
	} \\
		\subfloat[]{
        \includegraphics[width=.45\linewidth]{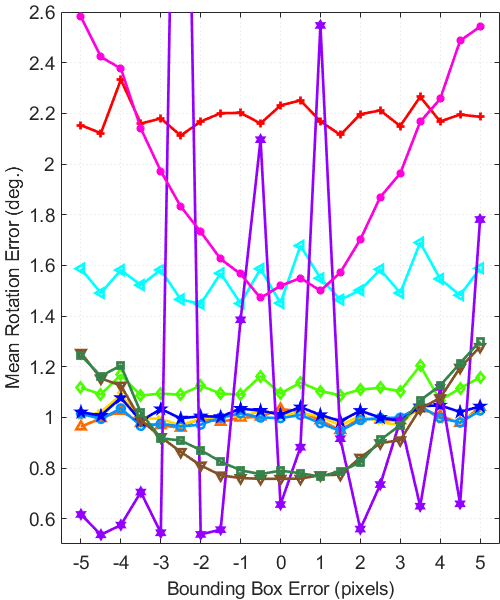}
	}
	\subfloat[]{
        \includegraphics[width=.45\linewidth]{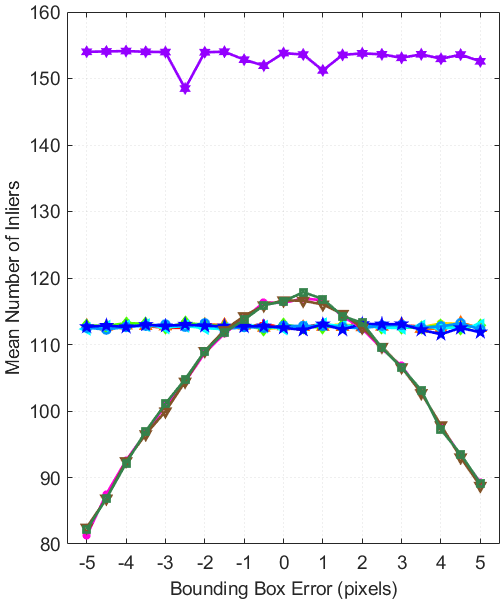}
	} 
	\end{center}
	\caption{High-quality images of Figs. 4(k)-4(n), i.e., the mean rotation errors and the number of inliers on \textit{E3} and \textit{E4}, respectively}
\end{figure*}

\end{document}